\documentclass[letterpaper]{article}
\usepackage{aaai22}
\usepackage{natbib}
\usepackage{times}
\usepackage{amssymb}
\usepackage{helvet}
\usepackage{courier}
\usepackage{amsfonts} 
\usepackage{booktabs}
\usepackage{graphicx}
\usepackage{float}
\usepackage{xspace}
\usepackage{xcolor}
\usepackage{amsmath}
\usepackage{dsfont}
\usepackage{subcaption}
\usepackage{paralist}
\usepackage{mathtools}
\usepackage{enumitem}
\usepackage{booktabs}
\usepackage{arydshln}
\usepackage{hyperref}
\usepackage{multicol,multirow}
\hypersetup{colorlinks=true,allcolors=violet}

\newcommand{\red}[1]{{\leavevmode\color{red}{#1}}}
\newcommand{\blue}[1]{{\leavevmode\color{blue}{#1}}}

\newcommand{\modelname}{{\sc GoalNet}\xspace}

\begin{document}
%
\title{
%
\modelname: Inferring Conjunctive Goal Predicates from Human Plan Demonstrations for Robot Instruction Following
%

%
%
}

\author{
    Shreya Sharma\textsuperscript{\rm 1 }\equalcontrib,
    Jigyasa Gupta\textsuperscript{\rm 1, \rm 2}\equalcontrib,
    Shreshth Tuli\textsuperscript{\rm 3},
    Rohan Paul\textsuperscript{\rm 1},
    Mausam\textsuperscript{\rm 1}\\
}
\affiliations{
    \textsuperscript{\rm 1}Indian Institute of Technology Delhi, India\\
    \textsuperscript{\rm 2}Samsung R\&D Institute India, Delhi\\
    \textsuperscript{\rm 3}Imperial College London, UK
}

\maketitle
\begin{abstract}
\begin{quote}
%
Our goal is to enable a robot to learn how to sequence its actions 
to perform tasks specified as natural language instructions, given successful demonstrations from a human partner. 
The ability to plan high-level tasks can be factored 
as (i) inferring specific goal predicates that characterize the task implied by a 
language instruction for a given world state and (ii) synthesizing a feasible goal-reaching action-sequence with such predicates.
For the former, we leverage a neural network prediction model, while utilizing a 
symbolic planner for the latter.
We introduce a novel neuro-symbolic model, \textsc{GoalNet}, for contextual 
and task dependent inference of goal predicates from human demonstrations 
and linguistic task descriptions. 
\textsc{GoalNet} combines (i) \emph{learning}, where dense representations
are acquired for language instruction and the world state that 
enables generalization to novel settings and (ii) \emph{planning},  
where the \textit{cause-effect} modeling by the symbolic planner eschews irrelevant predicates facilitating multi-stage decision making in large 
domains. 
 \textsc{GoalNet} demonstrates a significant improvement (51\%) in the task completion rate in comparison to a state-of-the-art rule-based approach on a benchmark data set displaying linguistic variations, particularly for multi-stage instructions.  
%

\end{quote}
\end{abstract}

\section{Introduction}

\begin{figure}[t]
    \centering \setlength{\belowcaptionskip}{-18pt}
    \setlength{\fboxsep}{1pt}%
    \fbox{\includegraphics[width=0.85\linewidth]{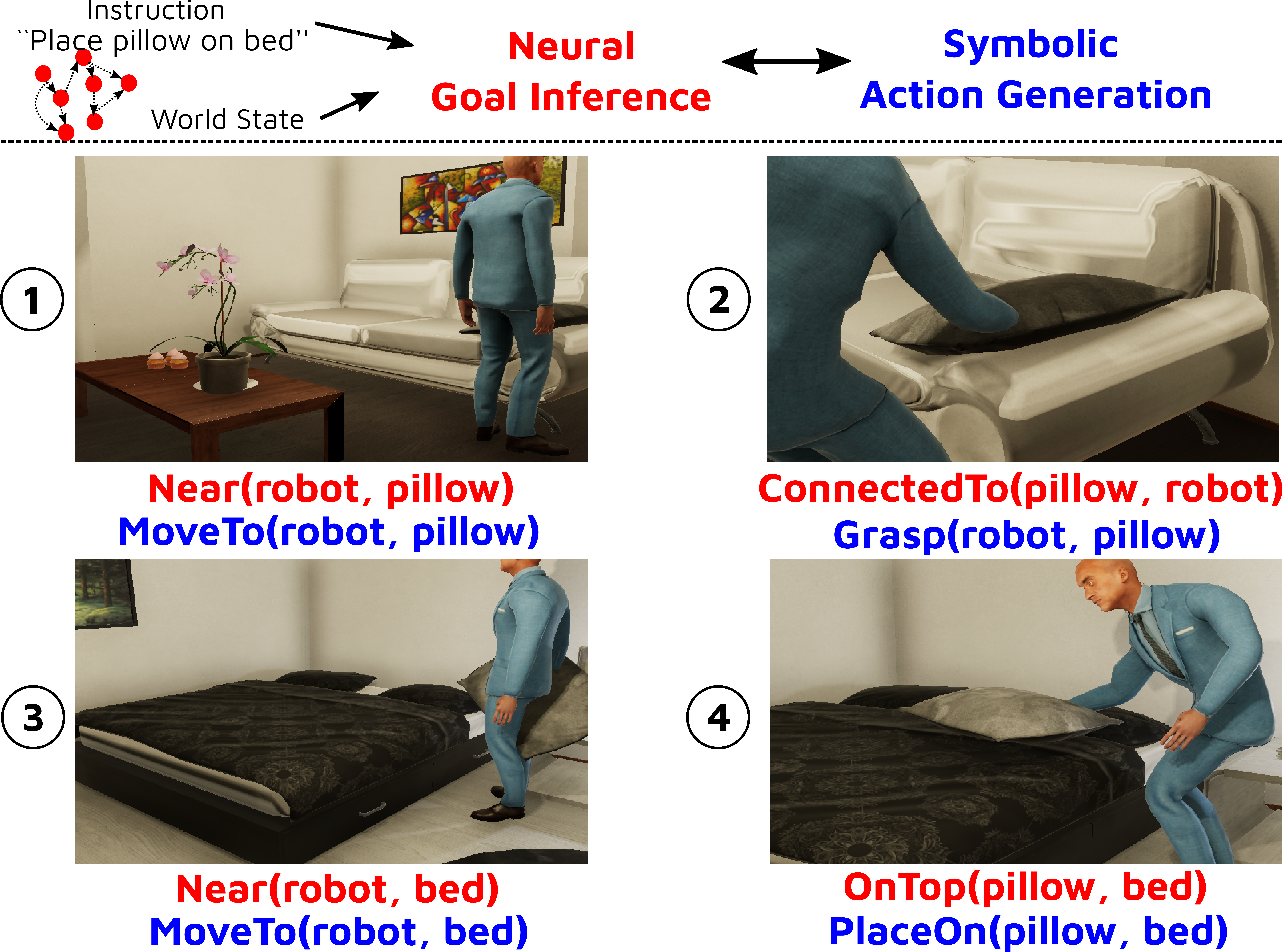}} \vspace{-8pt}
    \caption{\footnotesize We consider the task of inferring conjunctive goal predicates (red) from input world state and language instruction such that when given to a symbolic planner (blue), we reach a goal state. }
    \label{fig:intro}
\end{figure}

Robots are used in various scenarios where they interact with humans to perform tasks. Understanding natural language instructions, in such interactive settings, is crucial to effectively attain the human intended tasks. Recent efforts aim to directly map input state and language instructions to actions by inferring intended goals at the start to generate an action plan~\citep{tuli2021tango,toolnet,mei2016talk,suhr2018situated}. Such methods tend to lack the ability to interact dynamically with humans or scale with environment complexity~\citep{misra2018mapping}. Thus, we decompose the problem of generating an action sequence into the inference of task-specific goal predicates and subsequent generation of actions. However, learning to predict goal predicates is challenging due to following reasons. First, the hypothesis space of goal constraints may be large due to irrelevant or side-effects of other constraints. Second, high-level tasks may involve multi-stage plans requiring the learner to uncover the sub-goals characterizing the overall task from human demonstrations~\citep{pflueger2015multi}. Finally, the robot may encounter new instructions and world states necessitating conceptual generalization beyond limited training data. 

To make goal inference tractable in large-scale domains with multi-stage instructions, we leverage a symbolic planner with action models (in PDDL form) allowing us \emph{explain-away} side-effect or irrelevant predicates. To handle linguistic variations, we utilize a crowd-sourced dataset to train a neural model to infer conjunctive goal predicates for an input world state and instruction. We conjecture that viewing multiple task demonstrations enable an agent to infer the predicates/constraints needed to reach the goal. We assume that the human demonstrates plans for varied natural language instructions (e.g., \emph{``move all the fruits to the kitchen''}, \emph{``fill the cup with water''} etc.). The human demonstration constitute language instructions with state-action transitions leading to goal-reaching states. Due to the high variability in human language, learning through demonstrations facilitates goal reaching even with unseen instruction inputs. 

This paper addresses the problem of interpreting a task instruction from non-expert human users as a succinct set of goal predicates. For such users, we hypothesize that it is easier to demonstrate tasks in lieu of directly providing symbolic description.
We propose a \textit{neuro-symbolic model}, which we call \modelname, to infer goal predicates/constraints for a given state of the environment and natural language instruction. These constraints can then be passed to a high-level \textit{symbolic} planner to give out a sequence of actions in an interleaved fashion (see Fig.~\ref{fig:intro}).  This style of using a planner for forward simulation is unique to this work.
\modelname learns dense representations for instruction, world state and state history, enabling generalization to settings unseen during training. Our evaluation shows that the interactive dynamics between the neural model and the symbolic planner enables a robot to complete higher-level tasks such as cleaning a table, fetching fruits, arranging pillows, or even preparing meals in complex environments or scenes. Our results demonstrate a significant improvement of 51\% in the task completion rate in comparison to a state-of-the-art heuristic and rule-based approach for a mobile manipulator in a kitchen or living-room like environment with multi-stage instructions and linguistic variations.  

\section{Related Work}

\textbf{Grounding Instructions to Symbols. }  
 A line of work~\citep{thomason2015learning, howard2014natural} aims to train a supervised form of model that infers goal constraints for an input natural language instruction for classical instruction following. 
 Similarly,~\citet{squire2015grounding} ground language instructions to rewards functions. 
 Such annotations are challenging to provide and time-consuming for the human operator. 
 Instead, we adopt a framework in which only the natural demonstrations are provided to a  
 learner having access to the full environment state in lieu of explicit knowledge of 
 constraints corresponding to the input instructions.

 \textbf{Inferring plans from instructions.}  
 Alternatively, researchers have considered the problem of contextual induction of 
 multi-step plans from an instruction. 
 The seminal work of~\citet{misra2016tell} presents a probabilistic model that integrates 
 spatial cues and object properties in conjunction with a symbolic planner to 
 determine a grounded action sequence for an instruction. 
 In other work~\citep{misra2018mapping}, they attempt to increase the horizon of 
 plan induction by attempting to decouple the visual goal reasoning with motion planning. 
 In related work,~\citet{She2016IncrementalAO} present an incremental approach to interpret 
 verbs in the form of a ``constraint set'' hypothesis spaces that can be acquired 
 through an explicit feature-based representation. 
 We build on these works and make the following contributions. First, we 
 forego the need for hand-crafted features and learn a dense representation 
 for the environment and language enabling generalization. Second, 
 we cast the problem of goal constraint inference as a contextual sequence prediction 
 problem to generate larger sets of constraints. 
%
Alternatively, researchers cast the problem of instruction following as imitating 
tasks demonstrated by a human partner. 
Examples include,~\citep{tuli2021tango, liao2019synthesizing, bragg2020fake, shridhar2020alfred}.
The efforts aim at learning a policy that allows the robot or an embodied AI agent to 
complete tasks explicitly specified by a human in the form of a symbolic goal.  
%
%
This work address the complementary problem of inferring goal constraints and delegate the task of policy learning or planning to works as above.

\textbf{Goal Inference.} 
A growing body of work addresses the task of recognizing goals by observing the actions of agents and inferring a likelihood of each possible goal-predicate among the possible ones at each world state~\citep{meneguzzi2021survey}. 
The earliest work of \citet{lesh1995sound}  introduced a goal recognizer that observes human actions to prune inconsistent actions or goals from an input graph state. \cite{baker2009action} pose goal inference as probabilistic inverse planning and estimating a posterior distribution over sub-goals by observing human-generated plans. \cite{mann2021neural} presents an approach to infer goals even with sup-optimal or failed plans. 
Analogously, \cite{dragan2015effects} present a robot motion planner that generates legible plans using human-robot collaboration. \cite{boularias2015grounding} utilize inverse reinforcement learning to infer preference over predicates by using cost functions via human demonstrations.  
This paper addresses the related but different problem of inferring the specific hypothesis space that constitutes a goal for a task, taking a data-driven approach assuming the presence of \textit{human-optimal} demonstrations of the task in multiple contexts. The problem of disambiguation or attaining the intended goal is delegated to a planner or a learned policy.

\section{Problem Formulation}

\textbf{Robot and Environment Models.} We consider a robot as a autonomous agent with the ability to freely move and manipulate multiple objects in natural confined domains such as a \textit{kitchen} or \textit{living room} like environments. We consider objects as symbolic entities that consist of (i) identifying tokens such as ``apple'', ``stove'' and ``pillow'', (ii) object states such as $\mathrm{Open/Closed}$, $\mathrm{On/Off}$, and (iii) properties such as $\mathrm{isSurface}$, $\mathrm{isContainer}$, $\mathrm{isGraspable}$, etc. We consider object relations such as (i) \textit{support}: for example an apple supported on a tray or a shelf, (ii) \textit{containment}: for instance a pillow inside a box, carton or a cupboard, (iii) \textit{near}: an object being in close proximity to another, and (iii) \textit{grasped}: robot grasping a graspable object such as microwave door, fork, tap, etc. Let $s$ denote the state of a domain/environment in which an agent is expected to perform a task. The world state $s$ is a collection of symbolic objects including their identifiers, states and properties. The world state also includes object relations such as $\mathrm{OnTop}$, $\mathrm{Near}$, $\mathrm{Inside}$ and $\mathrm{ConnectedTo}$. We denote the set of spatial-relation types between objects and the set of object state constraints by $\mathcal{S}$. Examples include $\mathrm{OnTop(pillow_0, shelf_0)}$, $\mathrm{ConnectedTo(fork_0, robot)}$ and $\mathrm{stateIsOpen(tap_0)}$. Let $s_0$ denote the initial state of the domain and $\mathcal{O}(.)$ denote a map from an input state $s$ to a set of symbolic objects $\mathcal{O}(s)$ populating the world state $s$. Let $\mathcal{R}(.)$ denote a map from an input state $s$ to the set of object relations $\mathcal{R}(s)$. For any given state $s$ with relation set $\mathcal{R}(s)$, a relation $r \in \mathcal{R}(s)$ is denoted by $R(o^1, o^2)$ that represents a relation of type $R \in \mathcal{S}$ between objects $o^1 \in \mathcal{O}(s)$ and $o^2 \in \mathcal{O}(s)$ or as $R(o^1)$ in case of $r$ being an object state constraint. 

\textbf{Action and Transition Models.} We denote the set of all possible symbolic actions that the agent can perform on objects as $A$. For any given state $s$, an action $a \in A$ is represented in its abstract form as $I(o^1, o^2)$ where the interaction predicate $I \in \mathcal{I}$ affects the state or relation between $o^1$ and $o^2$. The interaction set $\mathcal{I}$ includes action predicates such as $\mathrm{Grasp}$, $\mathrm{MoveTo}$, $\mathrm{stateOn}$, $\mathrm{PlaceOn}$. Interaction effects are considered to be deterministic. For instance, a $\mathrm{Grasp}$ action establishes a \textit{grasped} relation between the agent and a $\mathrm{isGraspable}$ target object. A $\mathrm{stateOn}$ action applies to an $\mathrm{isTurnable}$ object, such as a tap, to swap its state between $\mathrm{On}$ and $\mathrm{Off}$. A $\mathrm{MoveTo(robot, couch_0)}$ action establishes a \textit{near} relation between the two objects $\mathrm{robot}$ and $\mathrm{couch_0}$. Similarly, a $\mathrm{PlaceOn(pillow_0, couch_0)}$ action establishes a \textit{support} relation from $\mathrm{pillow_0}$ and $\mathrm{couch_0}$. Interactions are also associated with pre-conditions in the form of relations or properties. For instance, $\mathrm{PlaceOn}$ and $\mathrm{PlaceIn}$ actions are allowed only when the object has a \textit{grasped} relation with the agent. Also, a $\mathrm{Grasp}$ action is permitted only when the target object has the $\mathrm{isGraspbable}$ property. For more details see Appendix~\ref{app:dataset_domain}. In our formulation, we assume presence of a low-level motion planner and delegate the problem of navigation and dexterous object manipulation to other works such as by~\citet{fitzgerald2021modelingref8, gajewski2019adaptingref9, lee2015learningref11}. We denote the deterministic transition function by $\mathcal{T}(\cdot)$. Thus, we can generate the successor state $s_{t+1} \gets \mathcal{T}(a_t, s_t)$ upon taking the action $a_{t}$ in state $s_{t}$.

\textbf{Tasks and Goals.} Given an initial state of the environment $s_0$ and transition model $\mathcal{T}$, the robot needs to perform a task in the form of a \textit{declarative} natural language instruction $l$. An instruction $l$ is encoded in the form of a sequence $\{ l_0, \ldots, l_z \}$ where each element is a token.  Each instruction $l$ corresponds to two sets $\Delta^+_l$ and $\Delta^-_l$, both being sets of symbolic goal constraints among world objects. The presence of $\Delta^+_l$ constraints and the absence of $\Delta^-_l$ constraints in the final state characterizes \textit{a} successful execution for the input instruction and are referred to as \textit{positive} and \textit{negative} constraints respectively. For example, for the input state $s$ with a pillow on the shelf and another inside a cupboard, the \emph{declarative} goal, $l = $ ``put the shelf pillow on the couch'' can be expressed as sets of constraints $\Delta^+_l = \{ \mathrm{OnTop(pillow_0, couch_{0})}\}$ and $\Delta^-_l = \{\mathrm{OnTop(pillow_0, shelf_0)}\}$. To successfully execute an input instruction $l$ from an initial state $s_0$, the agent must synthesize a plan as a sequence of actions $\{a_0, \ldots, a_T\}$ such that the final state $s_T = \mathcal{T}(\ldots \mathcal{T}(s_0, a_0) \ldots, a_T)$ consists of the goal constraints $\Delta^+_l$ and removes the constraints $\Delta^-_l$. Let $\mathcal{G}(s, l)$ denote the \emph{goal check} function that determines if the intended goal $l$ is achieved by a state $s$ as
\begin{equation}
    \setlength{\belowdisplayskip}{1pt} \setlength{\belowdisplayshortskip}{1pt}
    \setlength{\abovedisplayskip}{1pt} \setlength{\abovedisplayshortskip}{1pt}
    \mathcal{G}(s, l) = \mathds{1} \big( (\Delta^+_l \subseteq R) \wedge (\Delta^-_l \cap R = \emptyset) \big),
\end{equation}
where $R = \mathcal{R}(s)$ and $\Delta^+_l, \Delta^-_l$ are constraint sets for instruction $l$. We represent the set of all states that give $\mathcal{G}(s, l) = 1$ as $S^l$ and refer to them as goal states in further discussion. We denote the positive and negative constraints established at each step from state $s_t$ to $s_{t+1}$ by $\delta^+_t$ and $\delta^-_t$.  We also denote the \textit{constraint history} till time $t$ by $\eta_{t} = \{ (\delta^+_0, \delta^-_0), \dots, (\delta^+_{t-1}, \delta^-_{t-1}) \}$.

\textbf{Learning to Reach Goals.} This work aims to reach a goal state, \textit{i.e.} $\mathcal{G}(s_0, l) = 1$, given an initial world state $s_0$ and a natural language instruction $l$. We consider a discrete-time control problem of producing a goal-reaching plan by producing a policy that estimates an action sequence. Predicting actions for high-level multi-stage instructions can be partitioned into first predicting the required goal constraints to be achieved and then producing an action~\citep{misra2018mapping}. We focus on learning \textit{how} to infer goal-predicates and leverage a parameterized function $f_\theta(.)$ with parameters $\theta$ to determine the predicates to be achieved for the given state, constraint history and input instruction as 
\begin{equation}
    \delta^+_t, \delta^-_t = f_\theta(s_t, l, \eta_t).
\end{equation}
We also leverage a given planner $\mathcal{P}(.)$ that takes state, goal predicates and symbolic cause-effect domain definitions (denoted by $\Lambda$) as inputs and generates an action sequence $\mathcal{P}(s_t, \delta^+_t, \delta^-_t, \Lambda)$. Thus, at time-step $t$, our action $a_t$ is realized as $a_t = \mathcal{P}(s_t, f_\theta(s_t, l, \eta_t), \Lambda)$. 

Thus, we perform an interleaved predicate inference (using $f_\theta$) and feasible plan generation (using $\mathcal{P}$) at each time-step to reach a goal state. Formally, let $S^{\mathcal{P}, f_\theta, s}_{t}$ be a random variable denoting the state resulting from interleaved execution of planner $\mathcal{P}(.)$ and goal predictions $f_\theta(.)$ from state $s$ for $t$ time steps. We aim to learn $f_\theta(.)$ such that given to a planner $\mathcal{P}(.)$ the resultant state $S^{\mathcal{P}, f_\theta, s}_{k}$ reached in up to $T$ steps \textit{is a goal state} and the size of the set of inferred \textit{predicate set is minimized}. Thus, we have
\begin{equation*}
    \setlength{\belowdisplayskip}{1pt} \setlength{\belowdisplayshortskip}{1pt}
    \setlength{\abovedisplayskip}{1pt} \setlength{\abovedisplayshortskip}{1pt}
\begin{aligned}
\label{eq:problem}
& \underset{\theta}{\text{minimize}}
& & \|\delta^+_t \cup \delta^-_t\| \\
& \text{s. t.}
& & \forall\ t, \delta^+_t, \delta^-_t = f_\theta(s_t, l, \eta_t), \\
&&& \forall\ t, \mathcal{P}(s_t, \delta^+_t, \delta^-_t, \Lambda) \text{ is executed}.\\
&&& \exists\ k \in \{1, \ldots, T\}, \text{ s.t. } \mathcal{G}(S^{\mathcal{P}, f_\theta, s_0}_{k}, l) = 1.\\
\end{aligned}
\end{equation*}

Our function $f_{\theta}(.)$ is trained as a likelihood prediction model for the goal constraints leveraging a crowd-sourced dataset of task demonstrations. We denote a dataset $D_{\mathrm{Train}}$ of $N$ goal-reaching plans as
\begin{equation*}
    \setlength{\belowdisplayskip}{1pt} \setlength{\belowdisplayshortskip}{1pt}
    \setlength{\abovedisplayskip}{1pt} \setlength{\abovedisplayshortskip}{1pt}
    \mathcal{D_{\mathrm{Train}}} = \{ (s_0^i,l^i,\{s_j^i,a_j^i\}) \mid i \in \{1,N\}, j \in \{0,t_{i}-1\} \}, 
    \label{eq:data set}
\end{equation*}
where the $i^{th}$ datum  consists of the initial state $s^{i}_0$, the instruction $l^{i}$ and a state-action sequence $\{ (s^{i}_{0}, a^{i}_{0}), \dots, (s^{i}_{t-1}, a^{i}_{t-1}) \}$ of length $t_{i}$. We assume that human demonstrations reach the goal state and are optimal. The set of ground-truth constraints $s_t \setminus s_{t-1}$ and $s_{t-1} \setminus s_t$ become the supervision samples to train our $f_\theta(.)$ function as a neural model and generate learned parameters $\theta^*$. Akin to a typical \textit{divide-and-conquer} solution, this decomposition exemplifies our efforts to simplify the end-to-end problem of reaching goal states. 
\section{Technical Approach}
\label{sec:approach}

\begin{figure}[t]
    \centering \setlength{\belowcaptionskip}{-16pt}
    \includegraphics[width=\linewidth]{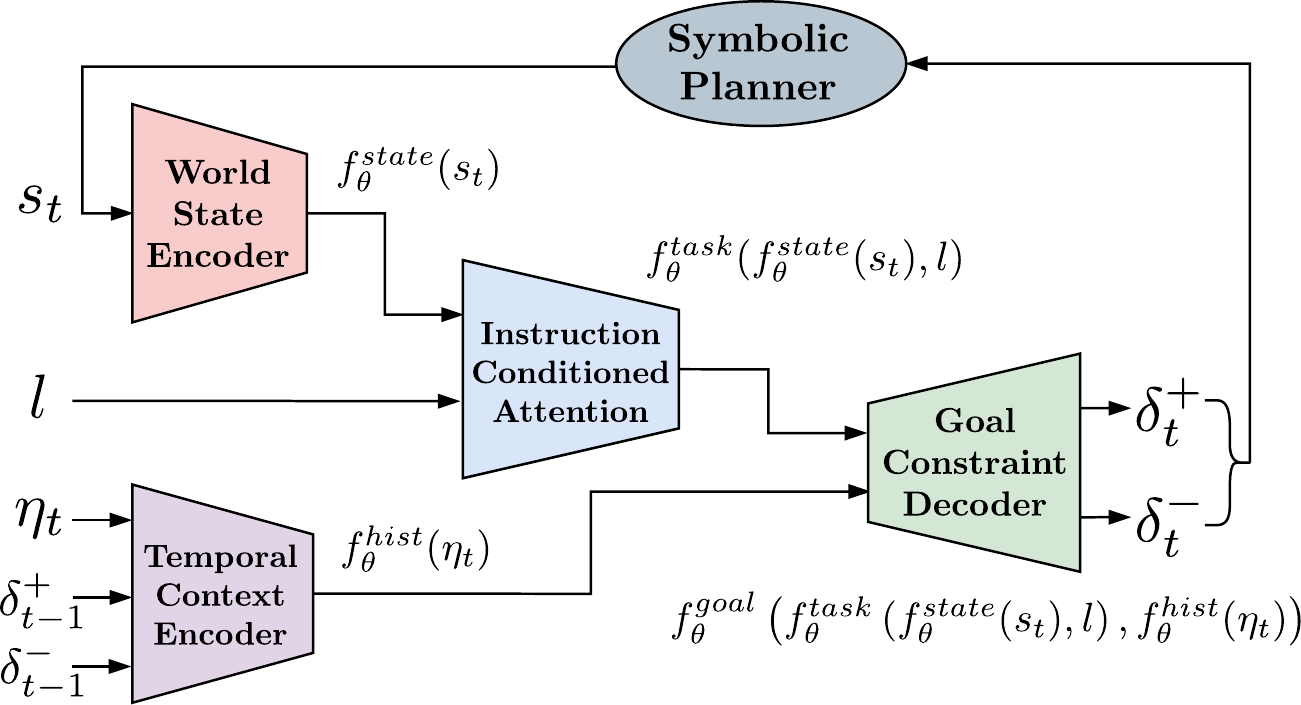}\vspace{-7pt}
    \caption{\footnotesize Using a symbolic planner and \modelname in tandem.}
    \label{fig:model}
\end{figure}

\begin{figure*}[t]
    \centering \setlength{\belowcaptionskip}{-15pt}
    \includegraphics[width=0.95\linewidth]{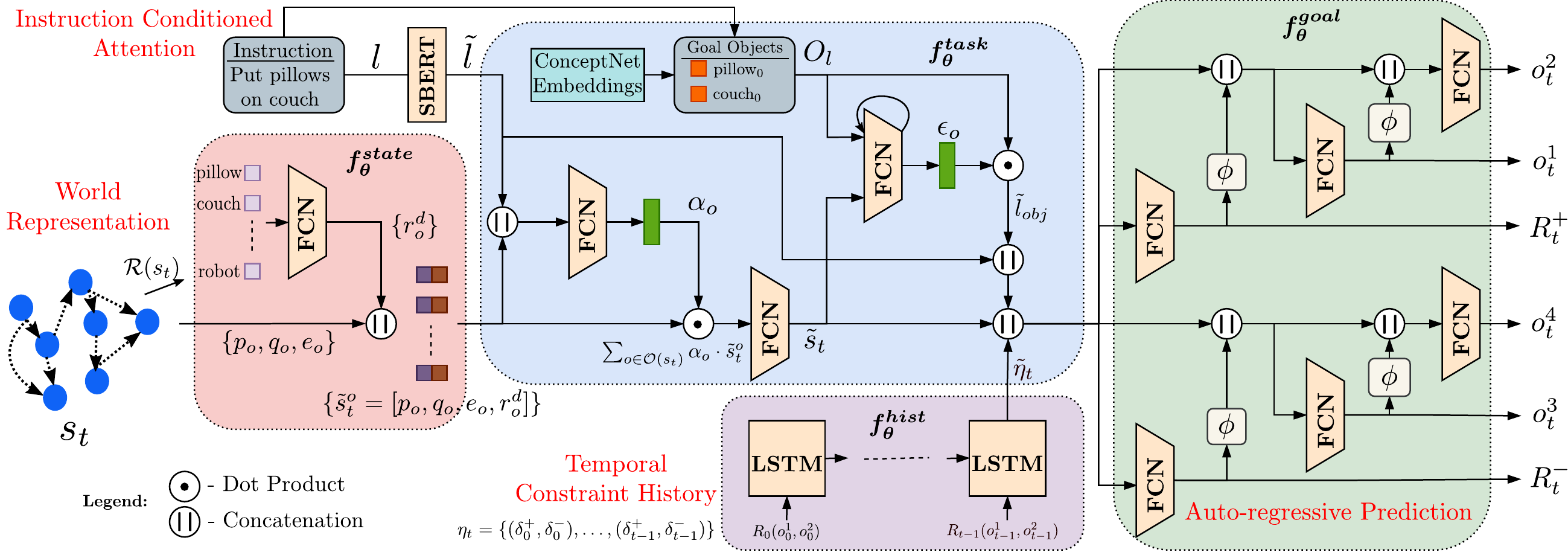}\vspace{-7pt}
    \caption{\footnotesize Details of the (colored) blocks from Figure~\ref{fig:model}. \modelname neural model encodes the world state and uses goal information with the natural language instruction input to attend over a task-specific context, finally decoding the next symbolic constraints to send to an underpinning planner.}
    \label{fig:nn}
\end{figure*}

We learn to predict the  next robot action $a_t$, given world state $s_t$, instruction $l$ and constraint history $\eta_t$. Assuming a given planner $\psi$, we realize the \modelname neural model for goal-constraint prediction as follows (see Fig.~\ref{fig:model}):
\begin{equation*}
\resizebox{\linewidth}{!}{
    $\delta^+_t, \delta^-_t = f_\theta(s_t, l, \eta_t) = f^{goal}_\theta \left( f^{task}_\theta \left( f^{state}_\theta(s_t), l \right), f^{hist}_\theta (\eta_t)  \right).$}
\end{equation*}
To do this, we encode the world state in the form of an object-centric graph. The state encoding is generated by fusing the relational and state information of the objects in the environment using the function $f^{state}_\theta(.)$. We encode the constraint history using the function $f^{hist}_\theta(.)$. We then attend over the state encoding conditioned on the input task instruction where the attention weights are generated via $f^{task}_\theta(.)$. Finally, the positive and negative constraints to be established at time step $t$ are predicted by the function $f^{goal}_\theta(.)$. The predicted constraints are then sent to the planner to generate an action $a_t$, executed in symbolic realizations of the environment and transition function and reach the next state $s_{t+1}$. We execute the above iteratively to reach a goal state. 

At each time step $t$, we encode the current world state $s_t$ as an object-centric graph $G_t = (\mathcal{O}(s_t), \mathcal{R}(s_t))$ where each node represents an object $o \in \mathcal{O}(s_t)$. Each relation $r \in \mathcal{R}(s_t)$ of the form $R(o^1, o^2)$ is encoded as a directed edge from $o^1$ to $o^2$ with edge type as $R$. To represent the states of an object $o$, we generate a binary vector $q_o = \{0, 1\}^u$ that represents the discrete object states for each of $u$ states that include $\mathrm{Open}/\mathrm{Closed}$, $\mathrm{On}/\mathrm{Off}$, etc. Similarly, we generate a binary vector $p_o = \{0, 1\}^v$ that represents the presence of various object properties (1 if present and 0 otherwise) for each of $v$ properties that include $\mathrm{isSurface}$ and $\mathrm{isContainer}$.  We also incorporate a function $\mathcal{C}(.)$ that generate a dense vector representation for an input token of an object. For an object $o$, we represent this by $e_o = \mathcal{C}(o) = \mathds{R}^w$ as a $w$-dimensional embedding. Our approach is agnostic to this function, but assumes that representations of semantically similar objects (such as with tokens ``apple'' and ``orange'') appear close, whereas semantically different objects appear far apart (such as with tokens ``fork'' and ``table'') in the learned space~\citep{mikolov2018advances}. Unless stated otherwise, we utilize $\mathrm{ConceptNet}$ embeddings~\citep{conceptnet-github} to facilitate generalization~\citep{tuli2021tango}. 

\noindent \textbf{World State Encoder.} We concatenate the embeddings $q_o$, $p_o$ and $e_o$ for each object $o \in \mathcal{O}(s_t)$ to form the object attributes that initialize each node of the graph representation of $s_t$. The relations of each object $o$ in the edges $\mathcal{R}(s_t)$ is represented as an adjacency vector $r_o$. The relational information is first encoded using a $d$-layer Fully Connected Network (FCN) with Parameterized ReLU (PReLU) \citep{prelu} activation as:
\begin{align}
\begin{split}
\label{eq:world_encoder}
    r_o^0 &= \mathrm{PReLU}( W^0_{rel}[r_o] + b^0_{rel} ), \\
    r_o^k &= \mathrm{PReLU}( W^k_{rel}[r_o^k] + b^k_{rel} ), \\
    r_o^d &= \mathrm{Sigmoid}( r_o^{d-1} ),
\end{split}
\end{align}
where $k$ varies from 0 to $d-1$, $W^k_{rel}$ and $b^k_{rel}$ represent the weight and bias parameters of the FCN. This results in a relational embedding for each object $o$ as $r_o^d$. Now, we fuse the semantic and relational embeddings to generate an embedding of each object $o$ as $[q_o, p_o, e_o, r_o^d]$. Late fusion of the relational information enables the downstream predictors to exploit the semantic and relational information independently, improving inference performance as demonstrated by prior work~\citep{toolnet}. Thus, the output of our state-encoder becomes
\begin{equation}
    f^{state}_{\theta}(s_t) = \{\tilde{s}_t^o\! =\! [p_o, q_o, e_o, r_o^d]|\ \forall o \in \mathcal{O}(s_t)\}.
\end{equation}

\noindent
\textbf{Temporal Context Encoder.} To ensure a seamless execution of the policy, it is crucial that the model is informed of the local context indicating the set of objects that the agent may manipulate in the future. This is typical in many real-life navigation and manipulation tasks where the sequential actions are temporally correlated. For instance, in the task of placing pillows on the couch, the agent first moves towards a pillow, grasps it and then places it on the couch. This entails that the sequence of goal constraints would initially have $\mathrm{Near(robot, pillow_0)}$, followed by $\mathrm{ConnectedTo(pillow_0, robot)}$ and then $\mathrm{OnTop(pillow_0, couch_0})$. This example demonstrates the high correlation between the interactions and manipulated objects of two adjacent time steps. Formally, \modelname encodes the temporal history of the goal-constraints $\eta_t$ using a Long-Short-Term-Memory (LSTM) neural model. For each relation $r = R_{t-1}(o^1_{t-1}, o^2_{t-1}) \in \mathcal{R}(s_{t-1})$ predicted in the previous time step $t-1$ in $\delta^+_{t-1} \cup \delta^-_{t-1}$, we define an encoding $\tilde{r} = [\vec{R}_{t-1}, \mathcal{C}(o^1_{t-1}), \mathcal{C}(o^2_{t-1})]$, where $\vec{R}_{t-1}$ is defined as a one-hot encoding for the relation type $R_{t-1} \in \mathcal{S}$ of the form $\{0, 1\}^{|S|}$.  $\mathcal{C}(o^1_{t-1})$ and $\mathcal{C}(o^2_{t-1})$ are the dense embeddings of the tokens of objects $o^1_{t-1}$ and $o^2_{t-1}$. At each time step $t$, we denote the encoding of the constraints history $\eta_t$ as $\tilde{\eta}_t$ where
\begin{equation*}
    f^{hist}_\theta = \tilde{\eta}_t = \mathrm{LSTM}([\vec{R}_{t-1}, \mathcal{C}(o^1_{t-1}), \mathcal{C}(o^2_{t-1})], \tilde{\eta}_{t-1}).
\end{equation*}

\noindent
\textbf{Instruction Conditioned Attention.} The input language instruction $l$ is encoded using a sentence embedding model represented as $\mathcal{B}(.)$ to generate the encoding $\tilde{l} = \mathcal{B}(l)$. We use the language instruction encoding $\tilde{l}$ to attend over the world objects using the following attention mechanism
\begin{align}
\begin{split}
    \alpha_o &= \mathrm{Sigmoid}( W_{attn} [ \tilde{s}_t^o, \tilde{l} ] + b_{attn} ),\ \forall o \in \mathcal{O}(s_t), \\
    \tilde{s}^l_t &= \sum_{o \in \mathcal{O}(s_t)} \alpha_o \cdot \tilde{s}_t^o,\\
    \tilde{s}_t &= \mathrm{PReLU}( W^k_{task}[\tilde{s}^l_t] + b^k_{task} )
\end{split}
\end{align}
Using a tokenizer~\citep{bird2009natural}, we also extract the set of objects in the instruction $l$ and denote this set as $O_l$. We then generate the encoding of the objects in the instruction using Bahdanau style state-conditioned self-attention~\citep{bahdanau2014neural} as
\begin{align}
\begin{split}
\label{eq:obj_attn}
    \epsilon_o &= \mathrm{Sigmoid}( W_l [\mathcal{C}(o), \tilde{s}_t] + b_l ),\ \forall o \in O_l,\\
    \tilde{l}_{obj} &= \sum_{o \in O_l} \epsilon_o \cdot \mathcal{C}(o).
\end{split}
\end{align}
This gives the output of $f^{task}_\theta(.)$ as
\begin{equation}
    f^{task}_\theta(\tilde{s}^o_t, l) = [\tilde{s}_t, \tilde{l}_{obj}, \tilde{l}].
\end{equation}
The attention operation aligns the information of the input language sentence with the scene to learn task-relevant context by allocating appropriate weights to objects. This relieves the downstream predictors from calculating the relative importance of world objects and focusing only on the ones related to the task, allowing the model to scale with the number of objects in the world.

\noindent
\textbf{Goal Constraint Decoder.} \modelname takes the instruction attended world state $\tilde{s}_t$, encoding of the constraint-history $\tilde{\eta}_t$, instruction objects encoding $\tilde{l}_{obj}$ and the sentence encoding $\tilde{l}$ to predict a pair of positive and negative constraints as relations $R_t^+(o^1_t, o^2_t)$ and $R_t^-(o^3_t, o^4_t)$. To predict each of the three components of relation and the two objects, we predict the likelihood score for each relation type $\mathcal{S}$ and objects $\mathcal{O}(s_t)$. We then select the relation or object with the highest likelihood scores. The three components are also predicted in an auto-regressive fashion. For instance, to predict $R_t^+(o^1_t, o^2_t)$, we first predict the relation $R_t^+$ and give the likelihood scores to the predictor for the first object $o^1_t$. Similarly, we forward likelihood scores of both $R_t^+$ and $o^1_t$ to predict object $o^2_t$. To do this, instead of using an $\mathrm{argmax}$ of the likelihood vector, we forward the Gumbel-Softmax of the vector~\citep{gumbel} (denoted by $\phi(.)$). It is a variation of $\mathrm{softmax}$ function that allows us to generate a one-hot vector while also allowing gradients to backpropagate (as $\mathrm{argmax}$ is not differentiable). It uses a temperature parameter $\tau$ that allows to control how close the output is to one-hot versus the $\mathrm{softmax}$ output. We use a small constant value as $\tau$ in our model. Thus, we generate the likelihood scores using the following mechanism
\begin{align*}
\begin{split}
    \tilde{R}_t^+ &= \mathrm{softmax}( W_R^+ [\tilde{s}_t, \tilde{\eta}_t, \tilde{l}_{obj}, \tilde{l}] + b_R^+ ),\\
    \tilde{o}_t^1 &= \mathrm{softmax}( W_1^+ [\tilde{s}_t, \tilde{\eta}_t, \tilde{l}_{obj}, \tilde{l}, \phi(\tilde{R}_t^+)] + b_1^+ ),\\
    \tilde{o}_t^2 &= \mathrm{softmax}( W_2^+ [\tilde{s}_t, \tilde{\eta}_t, \tilde{l}_{obj}, \tilde{l}, \phi(\tilde{R}_t^+), \phi(\tilde{o}_t^1)] + b_2^+ ).
\end{split}
\end{align*}
Then the predicate relation $R_t^+$ becomes $\mathrm{argmax}_{R \in \mathcal{S}} \tilde{R}_t^+$, the first object $o^1_t$ is $\mathrm{argmax}_{o \in \mathcal{O}(s_t)} \Omega (\tilde{o}_t^1, R_t^+)$ and $o^2_t$ as $\mathrm{argmax}_{o \in \mathcal{O}(s_t)} \Omega (\tilde{o}_t^2, R_t^+)$. A similar mechanism is followed to predict the negative constraint $R_t^-(o^3_t, o^4_t)$. Here, $\Omega$ denotes masks that impose grammar constraints at inference time based on pre-conditions that the relations impose. The masks are used to force the likelihood scores of infeasible objects to 0. For instance, the $\mathrm{OnTop}$ relation only accepts the objects as the second argument that have the $\mathrm{isSurface}$ property. We also mask out the likelihood scores of $\tilde{o}^2_t$ and $\tilde{o}^4_t$ based on whether $R^+_t$ and $R^-_t$ are constraints on the state of the objects in the first argument. To enable lexical action generation using a single positive and a single negative constraint, we use an anchored verb lexicon with post-conditions as described by~\citet{misra2015environment}. Thus, the predicted relations are represented as
\begin{align}
\begin{split}
    R_t^+(o^1_t, o^2_t), R_t^-(o^3_t, o^4_t) = f^{goal}_{\theta}(\tilde{s}_t, \tilde{\eta}_t, \tilde{l}_{obj}, \tilde{l}),\\
    \delta^+_t, \delta^-_t = \{R_t^+(o^1_t, o^2_t)\}, \{R_t^-(o^3_t, o^4_t)\}.
\end{split}
\end{align}
The model is trained using a loss \textit{viz} the sum of binary cross-entropy with ground-truth predicates for the six predictors.

\noindent \textbf{Symbolic Planner.} These constraint sets generated by the \modelname model are passed on to a planner $\psi$ that generates an action sequence as $\psi(s_t, \delta^+_t, \delta^-_t, \Lambda)$, that is executed by the agent. All pre-conditions and effects are encoded using a symbolic Planning Domain Definition Language (PDDL) file and is denoted as $\Lambda$. Unlike prior work that uses imitation or reinforcement learning~\citep{tuli2021tango, bae2019multi}, the PDDL information allows the planner to utilize symbolic information, indirectly informing the neural model. After running the planner, the executed action sequence and resulting world state is provided as input to the model for predicting the goal constraints and subsequently the action in the next step. Given such a planner, for every state $s^i_j$, we generate the goal-standard predicates using a single-step difference over the sets of relations between the two consecutive states as
\begin{gather*}
    \hat{\delta}^+_j, \hat{\delta}^-_j = s^i_{j+1} \setminus s^i_{j}, s^i_{j} \setminus s^i_{j+1} \forall j < t_i-1,\\
    \hat{\delta}^+_{t_i - 1}, \hat{\delta}^-_{t_i - 1} = \emptyset, \emptyset.
\end{gather*}
We encode the ground-truth set of predicates as binary vectors of relations and two objects. We then use the loss function to train the neural model in a teacher forced manner~\citep{toomarian1992learning}. However, at inference time, as we do not have ground-truth labels, we need to feed back the next state through emulation/planning, causing the model to be biased to the exposure and conditioning of the training data. To alleviate the effects of this bias, we use planners at two levels of fidelity: a \textit{low-fidelity} simulator and a \textit{high-fidelity} symbolic planner. In training, with a constant probability of $p$, we utilize a low-fidelity \textit{symbolic simulator} (referred to as $\textsc{SymSim}$ in the rest of the discussion) to train the model for each datapoint. \textsc{SymSim} emulates the effects of generating and executing action corresponding to the predicted goal predicates $\delta^+_t, \delta^-_t$ to generate the next state 
\begin{equation}
    \hat{s}^i_{j+1} = s^i_j \cup \delta^+_t \setminus \delta^-_t,
\end{equation}
where the $\cup$ and $\setminus$ operations are performed on the relation set of the graph $s^i_j$ to generate a new graph $\hat{s}^i_{j+1}$. In lieu of using a high-fidelity symbolic planner to generate actions and update the world state, this enables us to reduce training time significantly. In training, the recurrent predicate prediction stops when $\delta^+_t \cup \delta^-_t\! =\! \emptyset \vee t \geq t_i$, which we use as a proxy in lieu of explicitly learning the goal-check function.

On the other hand, at test time, we use a high-fidelity symbolic planner to generate the subsequent state as well as an action sequence at every time step $t$. The execution stops when $\delta^+_t \cup \delta^-_t = \emptyset \vee t \geq 30$, where the upper bound for recurrent execution is kept to thirty actions, \textit{viz}, the maximum plan length in the training dataset. The symbolic planner we use for evaluation is referred to as \textsc{Rintanen} and is taken from~\citet{RINTANEN201245}. This symbolic planner is able to predict actions and update the environment state with action effects and side effects, providing a complete update of the world state. We denote the updated state for an input datapoint with initial state $s^i_0$ as $\bar{s}^i_1$ and subsequent states as $\bar{s}^i_j$ where $j$ varies from 2 to $T_i$ with the length of the action sequence generated by \modelname is $T_i$. We similarly represent generated actions by $\bar{a}^i_j\ \forall j \in \{0, \ldots, T_i-1\}$. 

\noindent \textbf{Word and Sentence Embeddings.} 
\modelname uses word embedding function $\mathcal{C}(\cdot)$ that provides a  dense vector representation for word tokens associated with object class types. Word embeddings represent words meanings in a continuous space where vectors close to each other are semantically related, facilitating model generalization to novel object types encountered online. 
Using such (pre-trained) embeddings incorporates \emph{general purpose} object knowledge to facilitate richer generalization for downstream predicate learning. Similarly, we use $\mathrm{SentenceBERT}$ as our $\mathcal{B}(.)$ function that encodes the natural language instruction sentence~\citep{sbert}. This is a modification of the pre-trained BERT model~\citep{bert} that uses Siamese and triplet network structures to derive semantically meaningful sentence embeddings. The complete model is summarized in Fig.~\ref{fig:nn}. 



\section{Evaluation and Results}

\begin{table}[t]
    \centering \setlength{\belowcaptionskip}{-15pt}
    \resizebox{\linewidth}{!}{
    \begin{tabular}{@{}p{1.3\linewidth}@{}}
    \toprule
    \textbf{Objects:} (\textit{kitchen}) sink, stove, mug, microwave, fridge, icecream, kettle, coke,  plate, boiledegg, salt,  stovefire,  sinkknob. (\textit{living room}) loveseat, armchair, coffeetable, tv,  pillow, bagofchips, bowl, garbagebag, shelf, book, coke, beer.
    \tabularnewline
    \midrule
  \textbf{Object States:} $\mathrm{Open}/\mathrm{Closed}$, $\mathrm{On}/\mathrm{Off}$,  $\mathrm{HasWater}/\mathrm{HasChocolate}/\mathrm{IsEmpty}$, $\mathrm{DoorOpen}/\mathrm{DoorClosed}$. \tabularnewline
  \midrule
 \textbf{Object Properties:} $\mathrm{IsSurface}$, $\mathrm{IsTurnable}$, $\mathrm{IsGraspable}$, $\mathrm{IsPressable}$, $\mathrm{IsOpenable}$, $\mathrm{IsSqueezeable}$, $\mathrm{IsContainer}$. \tabularnewline
 \midrule
 \textbf{Actions:} $\mathrm{Grasp}$, $\mathrm{Release}$, $\mathrm{MoveTo}$, $\mathrm{PlaceOn}$, $\mathrm{PlaceIn}$, $\mathrm{Press}$, $\mathrm{Pour}$, $\mathrm{Squeeze}$, $\mathrm{stateOn}$, $\mathrm{stateOff}$, $\mathrm{stateOpen}$, $\mathrm{stateClose}$.\tabularnewline
 \midrule
    \textbf{Predicates:} $\mathrm{OnTop}$, $\mathrm{Near}$, $\mathrm{ConnectedTo}$. \tabularnewline
   \bottomrule
    \end{tabular}
    }\vspace{-7pt}
    \caption{\footnotesize Sample set of objects, states, properties, actions and predicates (for complete lists, see Table~\ref{tab:Objects_complete})}
    \label{tab:Objects}
\end{table}

\noindent \textbf{Data set Description.}
To demonstrate the efficacy of the \modelname model, we utilize the dataset made available by~\citet{misra2015environment}. The dataset consists of natural language instructions as a sequence of sentences, state, and action sequences collected through crowd-sourcing. The dataset is collected from two domains: kitchen and living room, each containing 40 objects, each with up to 4 instances of each object class. The dataset consists of diverse instruction types ranging from short-horizon tasks such as \emph{``go to the sink''} to high-level tasks involving complex and multiple interactions with environment objects such as \emph{``cook ramen in a pot of water"}. See Table~\ref{tab:Objects} for a list of sample objects in the domain, their possible states and properties with the set of symbolic actions allowed to be executed.

We extract each instruction sentence from a sequence of language instructions~\citep{She2016IncrementalAO}. We also extract the action sequence and environment states after each language instruction is executed in the environment. Thus, for each natural language instruction $l^i$, we have a state-action sequence $\{ (s^i_0, a^i_0), \ldots, (s^i_{t_i-1}, a^i_{t_i-1}) \}$ with the plan length of $t_i$. The dataset consists of 1117 such data points, where we use a split of $70\%: 15\%: 15\%$ datapoints for our train, validation and test datasets as per prior work~\cite{She2016IncrementalAO}. We also augment the training data to generate additional by perturbing the training data points by semantic object replacement in both states $\{s^i_j\}_j$ and actions $\{a^i_j\}_j$ as per the $\mathrm{ConceptNet}$ embeddings~\citep{tuli2021tango}. We perform this only for objects and datapoints in the training and validation sets wherein the replaced object is \textit{unseen} in the original data. This allowed us to increase the number of training datapoints by $25\%$, giving a total of 633 unique starting states (object-centric graphs) in the dataset, facilitating more diverse supervision to \modelname. The test set consists of initial states distinct from those in the training to ensure a fair and robust evaluation of predicate prediction. 

\noindent \textbf{Baseline and Evaluation Metrics.}
We adopt the approach by~\citet{She2016IncrementalAO} as the baseline. Building on the primitives developed by~\citet{misra2015environment}, this work maps closely to our setting where the objective maps directly to the prediction of goal predicates to be passed onto a symbolic planner. This approach addresses the problem of inferring a hypothesis space for verb frames extracted from language instructions. The determination of hypothesis space uses heuristics to focus the learner only on a focused set of predicates. Further, hand-crafted features extracted from language instruction, world state and the candidate goal states is used during learning. A log-linear model is then trained incrementally with the successful demonstration of a plan. We adopt this model as a baseline and compare the approach in a batch (instead of an incremental) setting. 

The accuracy of the goal-predicate predictions is evaluated as the ability to reach goal states using such predicates. We use popular metrics to evaluate the model~\citep{She2016IncrementalAO}.
We define the aggregate goal-predicates for a data point $\{ (s^i_0, a^i_0), \ldots, (s^i_{t_i-1}, a^i_{t_i-1}) \}$ and model generated sequence $\{ (s^i_0, \bar{a}^i_0), \ldots, (\bar{s}^i_{T_i-1}, \bar{a}^i_{T_i-1}) \}$ as
\begin{gather*}
\setlength{\belowdisplayskip}{0pt} \setlength{\belowdisplayshortskip}{0pt}
\setlength{\abovedisplayskip}{0pt} \setlength{\abovedisplayshortskip}{0pt}
    \hat{\Delta}^+_i, \hat{\Delta}^-_i = s^i_{t_i - 1} \setminus s^i_0, s^i_0 \setminus s^i_{t_i - 1},\\
    \Delta^+_i, \Delta^-_i = \bar{s}^i_{T_i - 1} \setminus s^i_0, s^i_0 \setminus \bar{s}^i_{T_i - 1}.
\end{gather*}
Thus, for a dataset of size $N$, we define
\begin{compactitem}
    \item \textit{SJI (State Jaccard Index)} checks the overlap between the aggregate predicates as 
    \begin{equation*}
    \setlength{\belowdisplayskip}{0pt} \setlength{\belowdisplayshortskip}{0pt}
    \setlength{\abovedisplayskip}{0pt} \setlength{\abovedisplayshortskip}{0pt}
        \text{SJI} = \frac{1}{N} \sum_{i = 1}^N \frac{\|\hat{\Delta}^+_i \cap \Delta^+_i \| + \|\hat{\Delta}^-_i \cap \Delta^-_i \|}{\|\hat{\Delta}^+_i \cup \Delta^+_i \| + \|\hat{\Delta}^-_i \cup \Delta^-_i \|} .
    \end{equation*}
    \item \textit{IED (Instruction Edit Distance):} measures the overlap between the similarity between the generated action sequence $\{\bar{a}^i_0, \ldots, \bar{a}^i_{T_i-1}\}$ and ground-truth sequence $\{a^i_0, a^i_{t_i-1}\}$. Specifically, the edit distance $d^i$ between these two sequences is used as
    \begin{equation*}
    \setlength{\belowdisplayskip}{0pt} \setlength{\belowdisplayshortskip}{0pt}
    \setlength{\abovedisplayskip}{0pt} \setlength{\abovedisplayshortskip}{0pt}
    \text{IED} = \frac{1}{N} \sum_{i = 1}^N 1 - \frac{d^i}{\max(T_i, t_i) }.
    \end{equation*}
    \item \textit{GRR (Goal Reaching Rate)} evaluates if the aggregated ground-truth predicates are present in the predicted ones
    \begin{equation*}
    \setlength{\belowdisplayskip}{0pt} \setlength{\belowdisplayshortskip}{0pt}
    \setlength{\abovedisplayskip}{0pt} \setlength{\abovedisplayshortskip}{0pt}
    \text{GRR} = \frac{1}{N} \sum_{i = 1}^N \mathds{1}(\Delta^+_i \subseteq \hat{\Delta}^+_i \wedge \Delta^-_i \subseteq \hat{\Delta}^-_i).
    \end{equation*}
    \item \textit{F1} evaluates the average of the F1 scores between the positive and negative aggregate predicate sets.  
\end{compactitem}

\begin{table}[t]
    \centering \setlength{\belowcaptionskip}{-15pt}
    \resizebox{\linewidth}{!}{
    \begin{tabular}{@{}lcccccc@{}}
    \toprule
    \textbf{Model} & \textbf{SJI} & \textbf{IED} & \textbf{F1} & \textbf{GRR}\tabularnewline
    \midrule
    Baseline \citep{She2016IncrementalAO} & 0.448  &  0.450 & 0.512 & 0.370 
 \tabularnewline
    \midrule
    \modelname &0.562  &0.601  &0.651  &0.468 \tabularnewline
    \midrule
    \textbf{Model Ablations} \tabularnewline
    \midrule
   - Relational information &0.533 &0.575 &0.621 &0.449\tabularnewline
   - Instance grounding &0.528 &0.567 &0.617 &0.439\tabularnewline
    - $\delta^-$ prediction &0.424  &0.447 &0.533 &0.416\tabularnewline
    - $\delta^+$ prediction &0.156 &0.167 &0.195 &0.098\tabularnewline
    - Temporal context encoding &0.221  &0.323  &0.263   &0.159 \tabularnewline
    - Goal Object Attn &0.547  &0.595   &0.634  &0.430 \tabularnewline
    - Instruction conditioned Attn &0.565  &0.604  &0.650  &0.456 \tabularnewline
    - Grammar mask &0.567  &0.602  &0.651  &0.459 \tabularnewline
    \midrule
    \textbf{Model Explorations} \tabularnewline
    \midrule
    Instruction encoding : Conceptnet &0.389  &0.451  &0.477  &0.297 \tabularnewline
    \midrule
    Temporal Context ($\delta^+_{t-1} \cup \delta^-_{t-1}$) &0.552 &0.580 &0.637 &0.430 \tabularnewline
    Temporal Context ($s_{t+1}$) &0.503 &0.553 &0.593 &0.380 \tabularnewline
    \midrule
    Training using  
    \textsc{Rintanen} &\textbf{0.645}  &\textbf{0.661}  &\textbf{0.717}  &\textbf{0.558} \tabularnewline
    \bottomrule
    \end{tabular}
    }
    \caption{\footnotesize A comparison of goal-prediction and goal-reaching performance for the baseline, the proposed \modelname model, ablations and explorations. Results are presented for test set derived from living room and kitchen domains.}
    \label{tab:Comparison with baselines and ablation}
\end{table}

\begin{figure*}
    \centering 
    \captionsetup[subfigure]{labelformat=empty}
    \begin{subfigure}[t]{0.196\linewidth}
        \centering
        \includegraphics[width=\linewidth, height=0.7\linewidth]{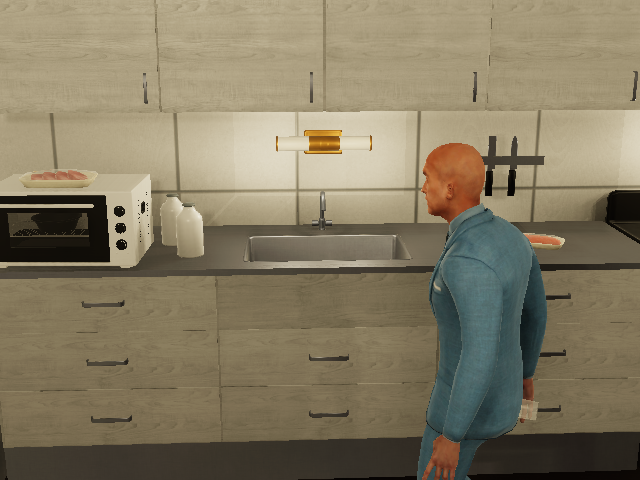} \vspace{-16pt}
        \caption{
        Initial State\\
        Instr.: \textit{``fill water in mug''}} \vspace{-6pt}
        \label{fig:fill_kettle_1}
    \end{subfigure}
    \begin{subfigure}[t]{0.196\linewidth}
        \centering
        \includegraphics[width=\linewidth, height=0.7\linewidth]{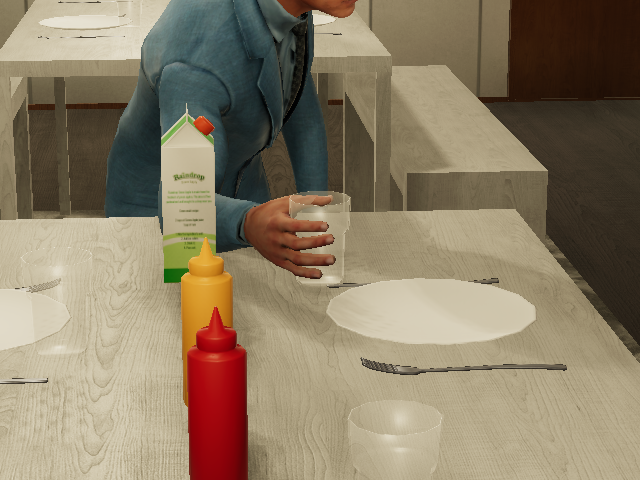} \vspace{-16pt}
        \caption{
        \scriptsize \red{$\mathrm{ConnectedTo(mug_0, robot)}$}}
        \scriptsize \blue{\{$\mathrm{MoveTo(robot, mug_0)}$, $\mathrm{Grasp(robot, mug_0)}$\}} \vspace{-6pt}
        \label{fig:fill_kettle_2}
    \end{subfigure}
    \begin{subfigure}[t]{0.196\linewidth}
        \centering
        \includegraphics[width=\linewidth, height=0.7\linewidth]{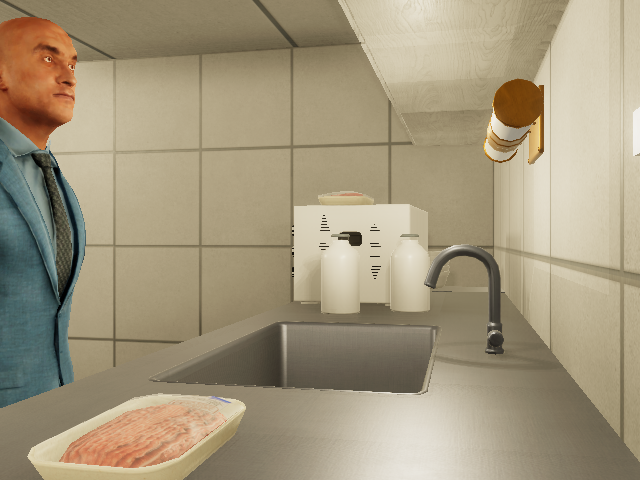} \vspace{-16pt}
        \caption{
        \scriptsize \red{$\mathrm{Near(robot, sink_0)}$}}
        \scriptsize \blue{\{$\mathrm{MoveTo(robot, sink_0)}$\}} \vspace{-6pt}
        \label{fig:fill_kettle_4}
    \end{subfigure}
    \begin{subfigure}[t]{0.196\linewidth}
        \centering
        \includegraphics[width=\linewidth, height=0.7\linewidth]{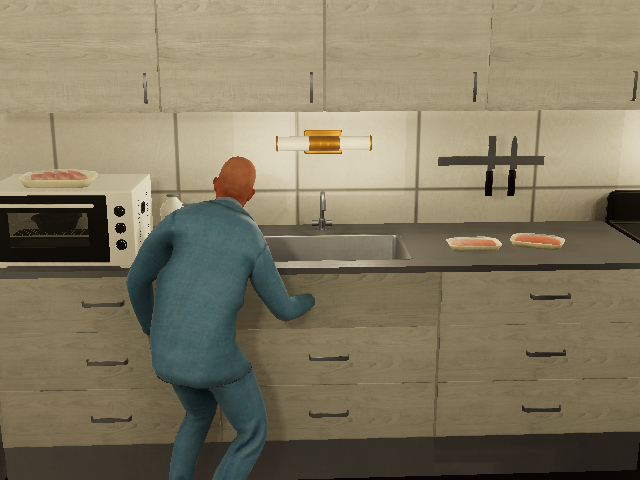} \vspace{-16pt}
        \caption{
        \scriptsize \red{$\mathrm{OnTop(mug_0, sink_0)}$}}
        \scriptsize \blue{\{$\mathrm{PlaceOn(mug_0, sink_0)}$\}} \vspace{-6pt}
        \label{fig:fill_kettle_3}
    \end{subfigure}
    \begin{subfigure}[t]{0.196\linewidth}
        \centering
        \includegraphics[width=\linewidth, height=0.7\linewidth]{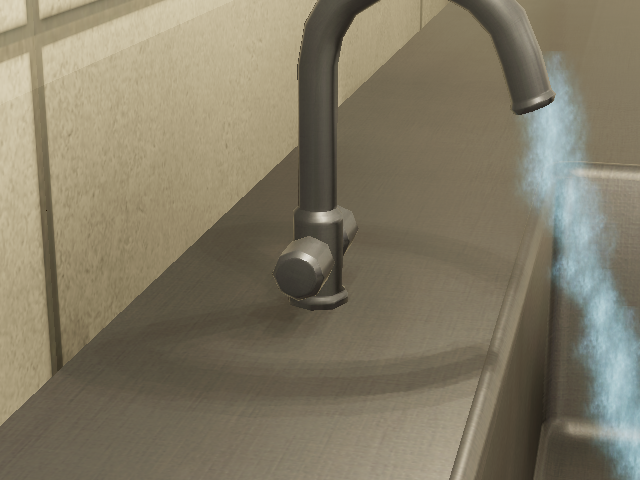} \vspace{-16pt}
        \caption{
        \scriptsize \red{$\mathrm{stateIsOn(tap_0)}$}}
        \scriptsize \blue{\{$\mathrm{stateOn(tap_0)}$\}} \vspace{-6pt}
        \label{fig:fill_kettle_5}
    \end{subfigure}
    \setlength{\belowcaptionskip}{-10pt}
    \caption{\footnotesize Sample plan in kitchen. Visualizations developed utilizing the VirtualHome Simulator~\citep{puig2018virtualhome} and a human-like agent with functionality akin to a single-arm manipulator. Predicted goal predicates shown in \red{red}. Executed plan at each time step shown in \blue{blue}. }
    \label{fig:fill_kettle}
\end{figure*}

\begin{figure*}
    \centering 
    \captionsetup[subfigure]{labelformat=empty}
    \begin{subfigure}[t]{0.196\linewidth}
        \centering
        \includegraphics[width=\linewidth, height=0.7\linewidth]{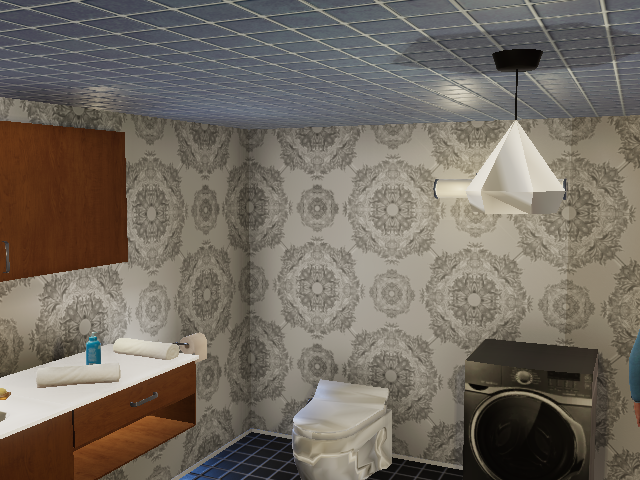} \vspace{-16pt}
        \caption{Initial State\\
        Intr.: \textit{``place beer and wine on top of the coffee-table''}} \vspace{-6pt}
        \label{fig:get_beer_and_wine_1}
    \end{subfigure}
    \begin{subfigure}[t]{0.196\linewidth}
        \centering
        \includegraphics[width=\linewidth, height=0.7\linewidth]{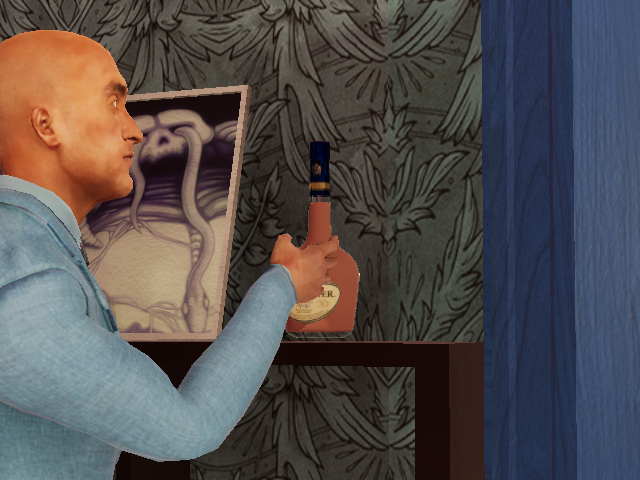} \vspace{-16pt}
        \caption{
        \scriptsize \red{$\mathrm{ConnectedTo(beer_0, robot)}$}}
        \scriptsize \blue{\{$\mathrm{MoveTo(robot, beer_0)}$, $\mathrm{Grasp(robot, beer_0)}$\}} \vspace{-6pt}
        \label{fig:get_beer_and_wine_2}
    \end{subfigure}
    \begin{subfigure}[t]{0.196\linewidth}
        \centering
        \includegraphics[width=\linewidth, height=0.7\linewidth]{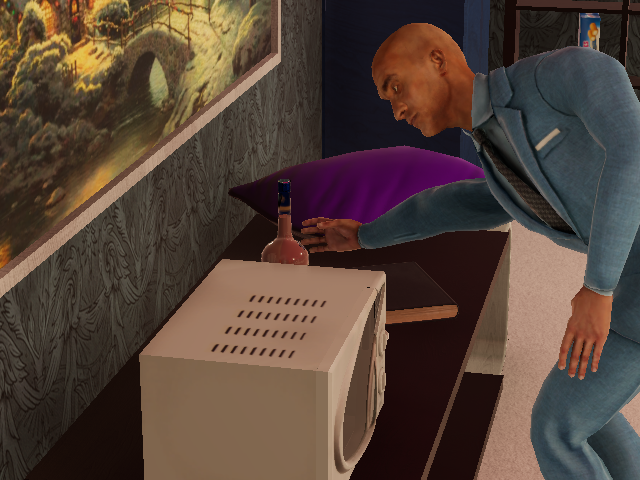} \vspace{-16pt}
        \caption{
        \scriptsize \red{$\mathrm{OnTop(beer_0, table_0)}$}}
        \scriptsize \blue{\{$\mathrm{MoveTo(robot, table_0)}$, $\mathrm{PlaceOn(beer_0, table_0)}$\}} \vspace{-6pt}
        \label{fig:get_beer_and_wine_3}
    \end{subfigure}
    \begin{subfigure}[t]{0.196\linewidth}
        \centering
        \includegraphics[width=\linewidth, height=0.7\linewidth]{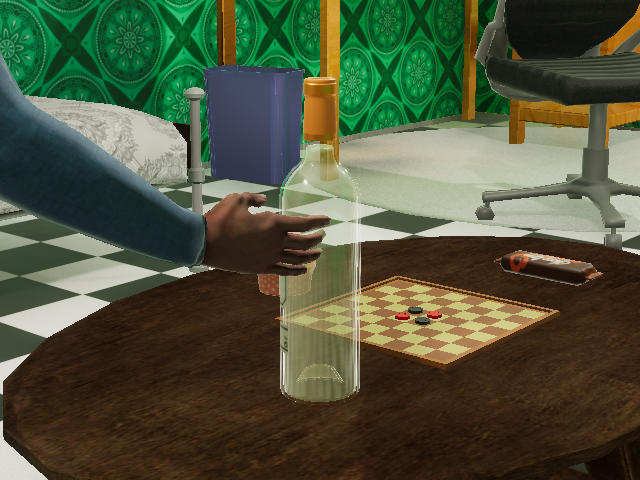} \vspace{-16pt}
        \caption{
        \scriptsize \red{$\mathrm{ConnectedTo(wine_0, robot)}$}}
        \scriptsize \blue{\{$\mathrm{MoveTo(robot, wine_0)}$, $\mathrm{Grasp(robot, wine_0)}$\}} \vspace{-6pt}
        \label{fig:get_beer_and_wine_4}
    \end{subfigure}
    \begin{subfigure}[t]{0.196\linewidth}
        \centering
        \includegraphics[width=\linewidth, height=0.7\linewidth]{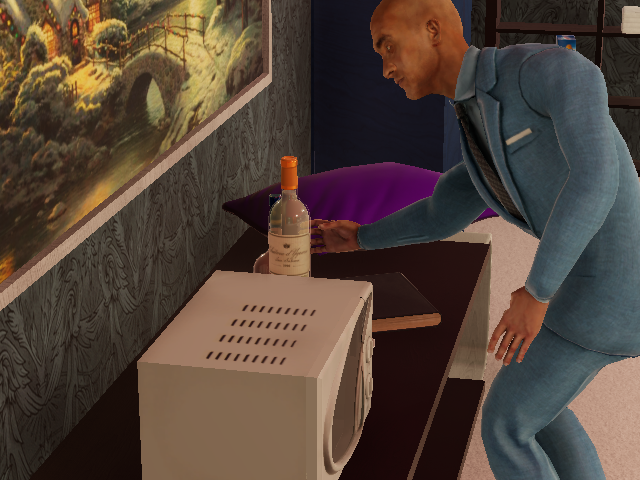} \vspace{-16pt}
        \caption{
        \scriptsize \red{$\mathrm{OnTop(wine_0, table_0)}$}}
        \scriptsize \blue{\{$\mathrm{MoveTo(robot, table_0)}$, $\mathrm{PlaceOn(wine_0, table_0)}$\}} \vspace{-6pt}
        \label{fig:get_beer_and_wine_5}
    \end{subfigure}
    \setlength{\belowcaptionskip}{-15pt}
    \caption{\footnotesize Sample plan in the living-room domain.}
    \label{fig:get_beer_and_wine}
\end{figure*}

\noindent \textbf{Baseline Comparison}
Table \ref{tab:Comparison with baselines and ablation} presents the scores for the baseline, \modelname and its variations. For a fair comparison with baseline, we have considered a grounding-aware version of \modelname where we include the instance groundings for all objects in the input instruction as part of the goal-object set $O_l$ (see Section~\ref{sec:approach}). \modelname improves SJI, IED, F1 and GRR by 25\%, 34\%, 27\% and 26\%, respectively. However, when we use \textsc{Rintanen} instead of \textsc{SymSim}, we get marked improvement, albeit taking $13\times$ more time to train the model. With \textsc{Rintanen} in the training loop, we get 44\%, 47\%, 40\% and 51\% higher SJI, IED, F1 and GRR scores demonstrating the importance of action side-effects introduced by the planner in lieu of using a simplified simulator. The improvement is primarily due to the dense representation of the input state, unlike the hand-crafted feature approach of~\citet{She2016IncrementalAO}, enabling \modelname to generalize to settings unseen at training. Figure \ref{fig:fill_kettle} and \ref{fig:get_beer_and_wine} shows state-action pair generated by \modelname in kitchen and living room domain respectively, demonstrating its ability to execute tasks successfully and reach a goal state.

\noindent \textbf{Analysis of Model Components}
Table \ref{tab:Comparison with baselines and ablation}  also presents scores corresponding to model ablations. We fix the model capacities for a fair comparison. 
Without the relational information in the form of adjacency matrix $\mathcal{R}(s_t)$ for input $s_t$, the model is unable to capture change in the spatial relations among world objects. For instance, when filling a mug with water (see Fig.~\ref{fig:fill_kettle}), the $\mathrm{PlaceOn(mug_0, sink_0)}$ establishes a $\mathrm{OnTop}$ relation between the two objects. Missing such changes in the state lead to a drop in performance of $\sim$ 5\%. 
Without the instance grounding of the objects the problem becomes harder. For instance, without knowing that \textit{coffee-table} in the instruction \textit{``place beer on top of coffee-table''} is mapped to $\mathrm{table_0}$, the agents needs to additionally infer the specific instance that needs to be manipulated in case multiple tables are present (see Fig.~\ref{fig:get_beer_and_wine}). We see a drop of $\sim$6\% in this case, though still higher than the baseline. 

An alternative approach could be to predict only positive or negative predicates. For instance, when we only predict positive predicates (\emph{- $\delta^-$ prediction}), GRR drops by $\sim$11\%. Such a model is unable to predict the required predicates when only relations are removed from the state. Examples include dropping objects when negative predicates include $\mathrm{Connected(robot, o)}$ for an object $\mathrm{o}$. Similarly, when only predicting the negative predicates (\emph{- $\delta^+$ prediction}), the model suffers a drop of $\sim$80\% in goal reaching performance. This evidence demonstrates the importance of predicting both positive and negative predicates to be sent to the underpinning planner/simulator. 
The inclusion of temporal context allows learning of correlated and commonly repeated action sequences. For instance, the task of filling a mug with water typically involves placing the mug beneath a faucet/tap and turning on the tap (see Fig.~\ref{fig:fill_kettle}). The ablation of this component leads to erroneous predictions when a particular predicate in a common plan fragment is missing or incorrectly predicted, for instance, when turning the tap on without placing the mug underneath.

Successful execution of an instruction may involve manipulation of multiple objects, such as beer and wine, when fetching both (see Fig.~\ref{fig:get_beer_and_wine}). Attention over the objects in the input language instruction (see eq.~\ref{eq:obj_attn}) enables \modelname to attend over the goal objects to dynamically prioritize manipulation at each time step. When we replace this (\emph{- Goal Object Attn}) with the mean of $\mathrm{ConceptNet}$ embeddings of goal objects, the GRR drops by $\sim$8\%. 
We use language instruction encoding $\tilde{l}$ to attend over the world objects. The attention operation aligns the information of the input language sentence with the scene to learn task-relevant context by allocating appropriate weights to objects. Without this, the GRR score drops by 2.6\%. 
Finally, without the grammar mask $\Omega$, the GRR drops marginally (2\%), showing the ability to learn grammar-related semantics of the domain.

\begin{table}[t]
    \footnotesize
    \centering
    \setlength{\tabcolsep}{4pt}
    \resizebox{\linewidth}{!}{
    \begin{tabular}{@{}lcccccccc@{}}
    \toprule 
    \multirow{2}{*}{\textbf{Model}} & \multicolumn{4}{c}{\textbf{Verb Replacement}} & \multicolumn{4}{c}{\textbf{Paraphrasing}}\tabularnewline
    \cmidrule{2-9} 
     & \textbf{SJI} & \textbf{IED} & \textbf{F1} & \textbf{GRR} & \textbf{SJI} & \textbf{IED} & \textbf{F1} & \textbf{GRR}\tabularnewline
    \midrule
    Baseline & 0.134  & 0.137 & 0.146 & 0.138 &  0.124 & 0.127 & 0.136  &  0.128 \tabularnewline
    GoalNet &  0.325 & 0.376  & 0.403 & 0.228 & 0.318 & 0.360 & 0.398 & 0.212 \tabularnewline
    \bottomrule
    \end{tabular}}
    \setlength{\belowcaptionskip}{-15pt}
     \caption{\footnotesize \modelname demonstrates the ability to generalize in case of \textit{verb replacement} and \textit{paraphrasing} relative to the baseline.}
    \label{tab:gen}
\end{table}

\noindent \textbf{Model Explorations.} 
We also explore additional variations of \modelname. For instance, when encoding the language instruction using $\mathrm{ConceptNet}$ instead of $\mathrm{SBERT}$, the scores drop by at least $\sim$26\%. This highlights the power of pre-trained language models in their ability to encode the task intention in natural language instructions. 
Additionally, we explore sending past predictions of both $\delta^+_t$ and $\delta^-_t$ predicates when encoding the temporal context. In such a case, the GRR score drops by 8\%. Similarly, when encoding temporal context using the encoding of the previous states, \textit{i.e.}, utilizing $\tilde{s}_t$, we see a drop in GRR by 19\%. This indicates that we need only the positive predicate information to encode the temporal context and additional information is superfluous to predict goal predicates effectively.

\noindent \textbf{Generalization.}
We additionally test the ability of \modelname to generalize to unseen instruction inputs by building two generalization data sets (see Table~\ref{tab:gen}). We test the performance when replacing verb frames in the training data with those absent in the data set. For instance, we replace \textit{boil} in \textit{``boil milk''} with \textit{heat}. Even in such cases, \modelname is able to successfully reach the goal state (see Fig.~\ref{fig:heat_milk}). We additionally paraphrase the language input to test instruction level generalization. For example, we paraphrase \textit{``gather all used plates and glasses, place into sink''} to \textit{``collect all used dishes and glasses, keep in wash basin''}. 

Table~\ref{tab:gen} above presents the performance scores of the baseline and \modelname models. It is observed that the baseline is unable to generalize to unseen verb frames and objects without any human intervention. On the other hand, \modelname generalizes in case of novel object references and unseen verbs relative to the baseline. \modelname improves GRR by 65\% in both cases of verb replacement and paraphrasing instructions. This generalization is achieved mainly due to the presence of dense token ($\mathrm{ConceptNet}$) and instruction ($\mathrm{SBERT}$) representations as opposed to storing observed verb-frame hypotheses in the baseline.

\section{Conclusions}
This paper proposes \modelname, a novel neural architecture that learns to infer goal predicates for an input language instruction and world state, which when passed to an underpinning symbolic planner enables reaching goal states. \modelname leverages independent inference over the objects in the world state and the spatial relations, applying instruction conditioned attention and using temporal contexts to autoregressively predict goal predicates. \modelname is trained using human demonstrations in kitchen and living-room environments and is able to generalize to unseen settings. 
This work demonstrates how learning and classical planning can be tied together to address the challenge of following multi-stage tasks for a robot. The neural model enables generalization to unseen language instructions, outperforming a state-of-the-art baseline in terms of the goal reaching performance. 
Future work will investigate out of domain generalization to apriori unknown number of objects, learning from sub-optimal or failed plans, and principled handling ambiguity among equally plausible goals.

\section*{Acknowledgments}
Mausam is supported by an IBM SUR award, grants by Google,Bloomberg and 1MG, Jai Gupta chair fellowship. Rohan Paul acknowledges
support from Pankaj Gupta Faculty Fellowship. Shreshth Tuli is supported
by the President’s Ph.D. scholarship at the Imperial College London. Jigyasa is supported by Samsung Research and Development Institute-Delhi.
We thank the IIT Delhi HPC facility for compute resources. We express
gratitude to Joyce Chai and Lanbo She for sharing code and dataset of \citet{She2016IncrementalAO}, Jussi Rintanen on his guidance to use SAT planner.

\bibliography{references}

\appendix


\section{Dataset and Domain details}
\label{app:dataset_domain}

\noindent \textbf{Domain Details.}  A typical world state of kitchen and living room environment consists of 40 objects each (Table \ref{tab:Objects_complete}). Some objects in the dataset do not play any functional role and have been removed to facilitate training. These include buttons of the Fridge and Microwave in the kitchen domain and buttons of the TV remote in the scenes from the living room domain.
Each object has an instance identifier ($\mathrm{mug_1,mug_2}$), a set of properties such as $\mathrm{IsGraspable}$ and $\mathrm{IsPourable}$ used for planning and a set of boolean states such as $\mathrm{HasWater/HasCoffee}$ that can be changed by robot actions. The robot
is also an object in the environment.
The environment model consist of 12 actions including $\mathrm{Grasp}$, $\mathrm{MoveTo}$ and $\mathrm{Pour}$, each having environment objects as arguments. Examples include $\mathrm{Grasp(robot, mug_1)}$ and $\mathrm{Pour(coke_0, mug_1)}$.
Each action results in effects encoded as postconditions in the domain description (via PDDL). The encoded effects take the form of conjunction of predicates or their negations. Action can introduce a new spatial relation between two objects, for example, $\mathrm{(OnTop(table_0,book_0)}$ or modify state of object, such as $\mathrm{State(kettle,HasWater)}$.

\begin{figure}[!t]
    \centering 
    \includegraphics[width=0.95\linewidth]{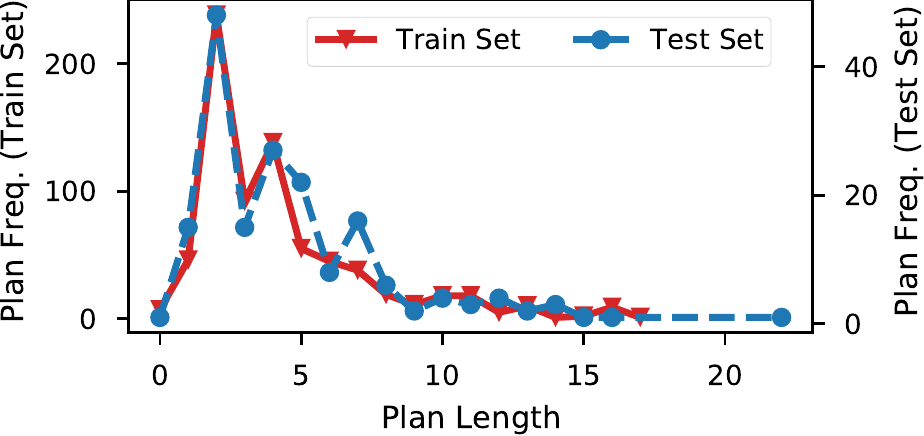}
    \caption{\footnotesize Frequency of plans with plan length.}
    \label{fig:freq}
\end{figure}

\begin{table*}[!t]
    \centering 
    \resizebox{\linewidth}{!}{
    \begin{tabular}{@{}p{1.1\linewidth}@{}}
    \toprule
    \textbf{Objects in Kitchen} :
    $\mathrm{robot}$,
    $\mathrm{ counter_1}$,
    $\mathrm{ sink}$,
    $\mathrm{ stove}$,
    $\mathrm{ mug_1}$,
    $\mathrm{ microwavedoor}$,
    $\mathrm{ microwave}$,
    $\mathrm{ fridge}$,
    $\mathrm{ fridgeleftdoor}$,
    $\mathrm{ fridgerightdoor}$,
    $\mathrm{ spoon_1}$,
    $\mathrm{ icecream_1}$,
    $\mathrm{ kettle}$,
    $\mathrm{ ramen_1}$,
    $\mathrm{ syrup_1}$,
    $\mathrm{ glass_1}$,
    $\mathrm{ longcup_1}$,
    $\mathrm{ longcup_2}$,
    $\mathrm{ fork_1}$,
    $\mathrm{ energydrink_1}$,
    $\mathrm{ coke_1}$,
    $\mathrm{ canadadry_1}$,
    $\mathrm{ plate_1}$,
    $\mathrm{ plate_2}$,
    $\mathrm{ syrup_2}$,
    $\mathrm{ instantramen_1}$,
    $\mathrm{ boiledegg_1}$,
    $\mathrm{ salt_1}$,
    $\mathrm{ light_1}$,
    $\mathrm{ stoveknob_1}$,
    $\mathrm{ stoveknob_2}$,
    $\mathrm{ stoveknob_3}$,
    $\mathrm{ stoveknob_4}$,
    $\mathrm{ stovefire_1}$,
    $\mathrm{ stovefire_2}$,
    $\mathrm{ stovefire_3}$,
    $\mathrm{ stovefire_4}$,
    $\mathrm{ fridgebutton}$,
    $\mathrm{ microwavebutton}$,
    $\mathrm{ sinkknob}$,
    $\mathrm{ icecreamscoop}$.
 \tabularnewline
    \midrule
    \textbf{Objects in Living Room}:
    $\mathrm{icecreamscoop}$,
   $\mathrm{robot}$,
   $\mathrm{ loveseat_1}$,
   $\mathrm{ armchair_1}$,
   $\mathrm{ armchair_2}$,
   $\mathrm{ coffeetable_1}$,
   $\mathrm{ tvtable_1}$,
   $\mathrm{ tv_1}$,
   $\mathrm{ tv_1remote_1}$,
   $\mathrm{ pillow_1}$,
   $\mathrm{ pillow_2}$,
   $\mathrm{ pillow_3}$,
   $\mathrm{ snacktable_1}$,
   $\mathrm{ bagofchips_1}$,
   $\mathrm{ bowl_1}$,
   $\mathrm{ garbagebag_1}$,
   $\mathrm{ garbagebin_1}$,
   $\mathrm{ shelf_1}$,
   $\mathrm{ shelf_2}$,
   $\mathrm{ book_1}$,
   $\mathrm{ book_2}$,
   $\mathrm{ coke_1}$,
   $\mathrm{ beer_1}$,
   $\mathrm{ xboxcontroller_1}$,
   $\mathrm{ xbox_1}$,
   $\mathrm{ cd_2}$,
   $\mathrm{ cd_1}$,
   $\mathrm{ tv_1powerbutton}$,
   $\mathrm{ tv_1channelupbutton}$,
   $\mathrm{ tv_1channeldownbutton}$,
   $\mathrm{ tv_1volumeupbutton}$,
   $\mathrm{ tv_1volumedownbutton}$,
   $\mathrm{ tv_1remote_1powerbutton}$,
   $\mathrm{ tv_1remote_1channelupbutton}$,
   $\mathrm{ tv_1remote_1channeldownbutton}$,
   $\mathrm{ tv_1remote_1volumeupbutton}$,
   $\mathrm{ tv_1remote_1volumedownbutton}$,
   $\mathrm{ tv_1remote_1mutebutton}$.
    \tabularnewline
    \midrule
  \textbf{Object States:}
  $\mathrm{Open}/\mathrm{Closed}$,
  $\mathrm{On}/\mathrm{Off}$,
  $\mathrm{DoorOpen}/\mathrm{DoorClosed}$,
  $\mathrm{HasIceCream}/\mathrm{HasEgg}/\mathrm{HasRamen}/\mathrm{HasCoffee}/\mathrm{HasCD}/\mathrm{HasWater}$,
  $\mathrm{HasChips},\mathrm{HasSpoon}/\mathrm{HasVanilla}/\mathrm{HasChocolate}/\mathrm{IsEmpty}/\mathrm{IsScoopsLeft}$, $\mathrm{OnChannel1}/\mathrm{OnChannel2}/\mathrm{OnChannel3}/\mathrm{OnChannel4}$, $\mathrm{VolumeUp/VolumneDown}$.\tabularnewline
  \midrule
 \textbf{Object Properties:} 
 $\mathrm{IsAddable}$, $\mathrm{IsScoopable}$,
 $\mathrm{isSurface}$, $\mathrm{IsGraspable}$, $\mathrm{isPressable}$, $\mathrm{IsTurnable}$, $\mathrm{IsSqueezeable}$, $\mathrm{isContainer}$,$\mathrm{isOpenable}$  \tabularnewline
 \midrule
 \textbf{Actions:} $\mathrm{Grasp}$, $\mathrm{Release}$, $\mathrm{MoveTo}$, $\mathrm{PlaceOn}$, $\mathrm{PlaceIn}$, $\mathrm{Press}$, $\mathrm{Pour}$, $\mathrm{Squeeze}$, $\mathrm{stateOn}$, $\mathrm{stateOff}$, $\mathrm{stateOpen}$, $\mathrm{stateClose}$.\tabularnewline
 \midrule
    \textbf{Predicates:} $\mathrm{OnTop}$, $\mathrm{Inside}$, $\mathrm{Near}$, $\mathrm{ConnectedTo}$. \tabularnewline
\midrule
\textbf{Action Preconditions}\tabularnewline
$\mathrm{Grasp(robot,o)} : \mathrm{IsGraspable(o) \wedge Near(robot,o) \wedge \neg ConnectedTo(o,robot)}$ \tabularnewline
$\mathrm{Release(robot,o)}:\mathrm{ConnectedTo(o,robot)}$ \tabularnewline
$\mathrm{MoveTo(robot,o)} : \emptyset$ \tabularnewline
$\mathrm{PlaceOn(o^1,o^2)} : \mathrm{ConnectedTo(o^1,robot) \wedge isSurface(o^2)}$ \tabularnewline
$\mathrm{PlaceIn(o^1,o^2)} : \mathrm{ConnectedTo(o^1,robot) \wedge Near(o^2,robot) \wedge isContainer(o^1)}$  \tabularnewline 
$\mathrm{Press(o^1)}: \mathrm{Near(robot,o^1) \wedge isPressable(o^1) }$  \tabularnewline
$\mathrm{Pour(o^1,o^2)}: \mathrm{ConnectedTo(o^1,robot) \wedge Near(o^2,robot)}$ \tabularnewline
$\mathrm{Squeeze(o^1,o^2)}: \mathrm{ConnectedTo(o^1) \wedge Near(o^2,robot) \wedge IsSqueezable(o^1)}$ \tabularnewline
$\mathrm{StateOn(o)}: \mathrm{Off(o) \wedge IsTurnable(o)}$ \tabularnewline
$\mathrm{StateOff(o)}: \mathrm{On(o) \wedge IsTurnable(o)}$ \tabularnewline
$\mathrm{StateOpen(o)}: \mathrm{Closed(o) \wedge IsOpenable(o)}$ \tabularnewline
$\mathrm{StateClose(o)}: \mathrm{Open(o) \wedge IsOpenable(o)}$ \tabularnewline
\midrule
\textbf{Actions Post-conditions}\tabularnewline
$\mathrm{Grasp(robot,o)} : \mathrm{ConnectedTo(o,robot)}$ \tabularnewline
$\mathrm{Release(robot,o)}: \mathrm{\neg ConnectedTo(o,robot)}$ \tabularnewline
$\mathrm{MoveTo(robot,o)} : \mathrm{Near(o,robot)}$ \tabularnewline
$\mathrm{PlaceOn(o^1,o^2)} : \mathrm{OnTop(o^1, o^2)}$ \tabularnewline
$\mathrm{PlaceIn(o^1,o^2)} : \mathrm{Inside(o^1, o^2)} $  \tabularnewline 
$\mathrm{Press(o^1)}: \mathrm{On(o^1)}$  \tabularnewline
$\mathrm{Pour(o^1,o^2)}: \mathrm{Inside(o^1, o^2)}$ \tabularnewline
$\mathrm{Squeeze(o^1,o^2)}: \mathrm{\neg HasWater(o^1) \wedge \neg Inside(o^1, o^2) } $ \tabularnewline
$\mathrm{StateOn(o)}: \mathrm{On(o)}$ \tabularnewline
$\mathrm{StateOff(o)}: \mathrm{Off(o)}$ \tabularnewline
$\mathrm{StateOpen(o)}: \mathrm{Open(o)}$ \tabularnewline
$\mathrm{StateClose(o)}: \mathrm{Closed(o)}$ \tabularnewline
   \bottomrule
    \end{tabular}
    }
    \caption{\footnotesize Complete list of objects in kitchen and living room domains. Properties and states of objects. Predicates between objects. Actions with their pre-conditions and post-conditions.}
    \label{tab:Objects_complete}
\end{table*}

\noindent \textbf{Dataset Details.}
The dataset is devoid of the metric information such as position as well as the point-cloud models that includes the object geometries, bounding boxes and pose data.
Both training and test datasets consist of long-horizon plans with up to 30 action sequences, albeit with decreasing frequency as we increase plan length (see Fig.~\ref{fig:freq}).
The original dataset of \citet{misra2015environment} defines ten high-level objectives, five for each domain. These include \textit{``make ramen'', ``clean the room'', ``make coffee'', ``prepare room for movie night''}, etc. Each high level task is described as a sequence of low level instructions mapped to an initial environment and action sequence. For instance, high level task of \textit{``make ramen''} is decomposed as \textit{``Take pot on counter and fill it with water from the sink''. ``Place the pot on a burner on the stove''.  ``Turn on the burner and let the water boil''. ``Once it is boiling,   open the lid of the ramen cup and pour boiling water in until it reaches the top of the container''}. \citet{She2016IncrementalAO} extract these low level instructions, each with a single verb and its arguments. 
The above decomposition leads to a training set with 77 unique verbs, an average of 6 words per instruction text and average plan length for train and test instances of $\sim$5. Out of 1117 data instances, 56\% are from kitchen domain and remaining from living room. We perform data cleaning as performed by~\citet{She2016IncrementalAO} to remove noise in the dataset, for instance, removing \textit{wait} statements in the discrete-time control setup we consider.  


\begin{figure*}[t]
    \centering
    \includegraphics[width=0.22\linewidth]{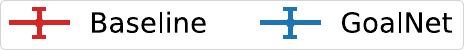}\\
    \begin{subfigure}{0.24\linewidth}
        \centering
        \includegraphics[width=\linewidth]{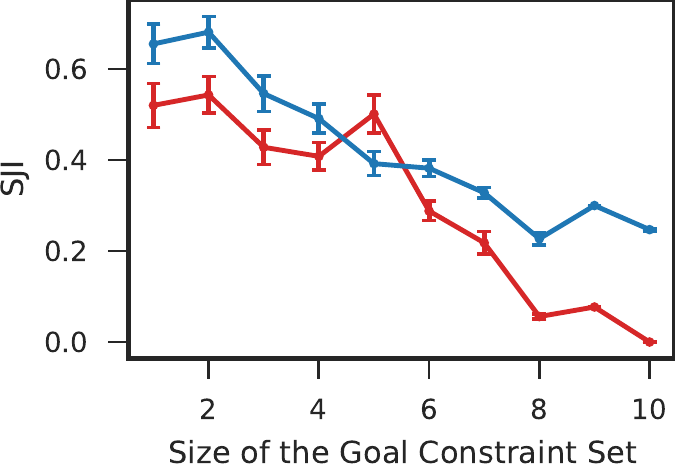}
        \caption{SJI Score} \vspace{-6pt}
        \label{fig:sens_sji}
    \end{subfigure}\ \ 
    \begin{subfigure}{0.24\linewidth}
        \centering
        \includegraphics[width=\linewidth]{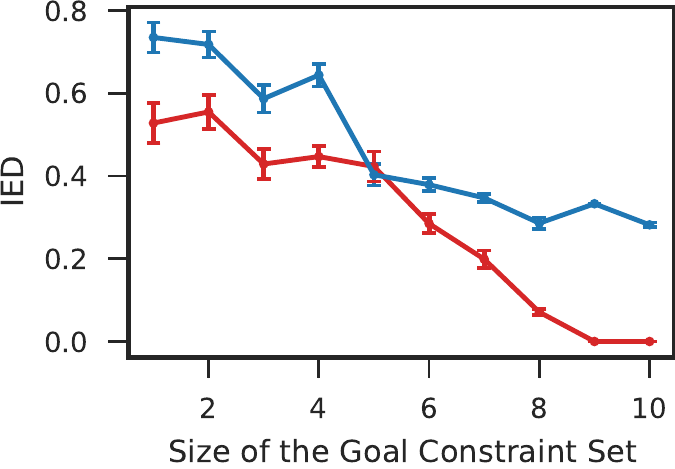}
        \caption{IED Score} \vspace{-6pt}
        \label{fig:sens_ied}
    \end{subfigure}\ \ 
    \begin{subfigure}{0.24\linewidth}
        \centering
        \includegraphics[width=\linewidth]{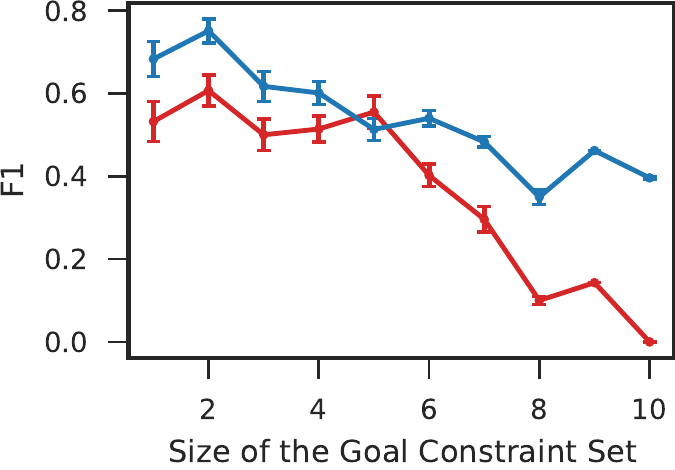}
        \caption{F1 Score} \vspace{-6pt}
        \label{fig:sens_f1}
    \end{subfigure}\ \ 
    \begin{subfigure}{0.24\linewidth}
        \centering
        \includegraphics[width=\linewidth]{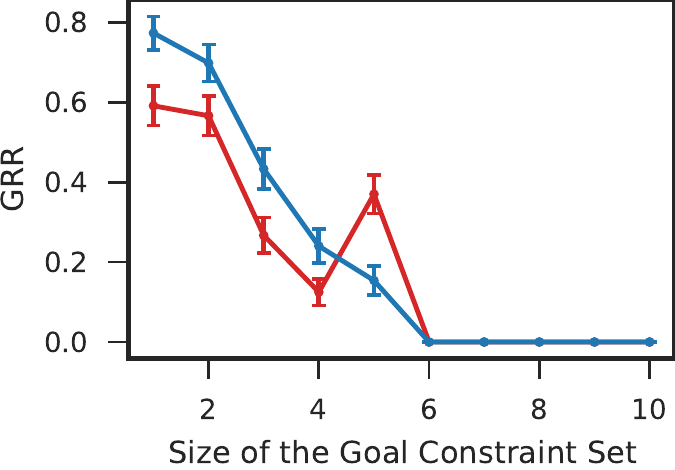}
        \caption{GRR Score} \vspace{-6pt}
        \label{fig:sens_grr}
    \end{subfigure}
    \caption{Performance of baseline and \modelname model with the size of aggregate goal-constraint sets.}
    \label{fig:sensitivity}
\end{figure*}

\begin{figure*}[t]
    \centering 
    \captionsetup[subfigure]{labelformat=empty}
    \begin{subfigure}[t]{0.196\linewidth}
        \centering
        \includegraphics[width=\linewidth, height=0.7\linewidth]{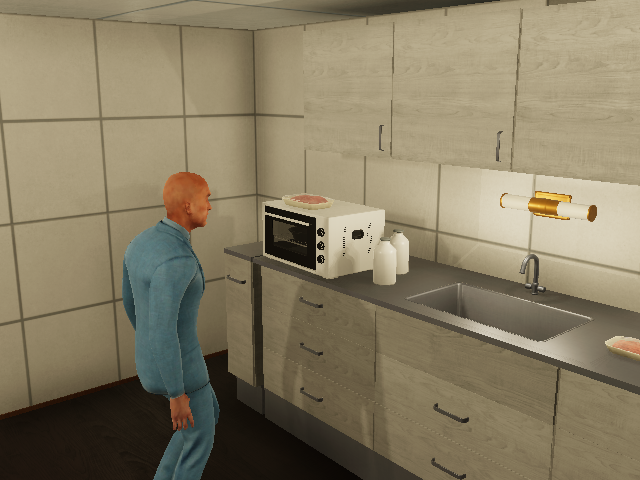} \vspace{-16pt}
        \caption{Initial State\\Intr.: \textit{``heat milk''}} \vspace{-6pt}
        \label{fig:heat_milk_1}
    \end{subfigure}
    \begin{subfigure}[t]{0.196\linewidth}
        \centering
        \includegraphics[width=\linewidth, height=0.7\linewidth]{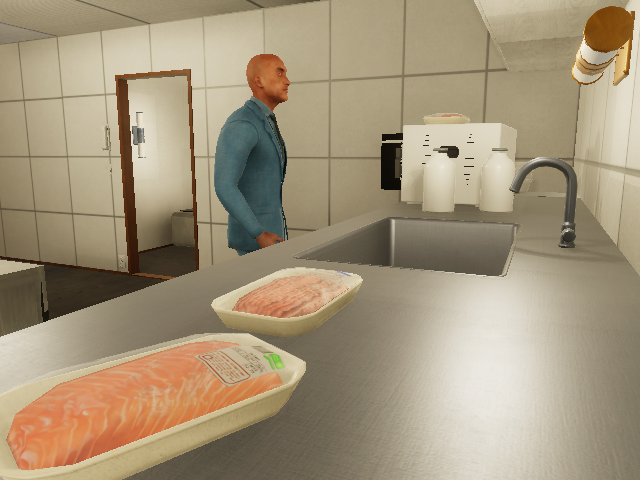} \vspace{-16pt}
        \caption{
        \scriptsize \red{$\mathrm{stateIsOpen(microwave_0)}$}}
        \scriptsize \blue{\{$\mathrm{MoveTo(robot, microwave_0)}$, $\mathrm{stateOpen(microwave_0)}$\}} \vspace{-6pt}
        \label{fig:heat_milk_2}
    \end{subfigure}
    \begin{subfigure}[t]{0.196\linewidth}
        \centering
        \includegraphics[width=\linewidth, height=0.7\linewidth]{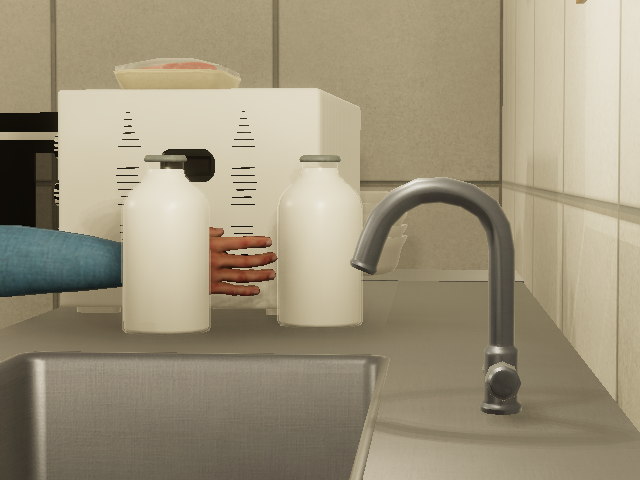} \vspace{-16pt}
        \caption{
        \scriptsize \red{$\mathrm{ConnectedTo(milk_0, robot)}$}}
        \scriptsize \blue{\{$\mathrm{MoveTo(robot, milk_0)}$, $\mathrm{Grasp(robot, microwave_0)}$\}} \vspace{-6pt}
        \label{fig:heat_milk_3}
    \end{subfigure}
    \begin{subfigure}[t]{0.196\linewidth}
        \centering
        \includegraphics[width=\linewidth, height=0.7\linewidth]{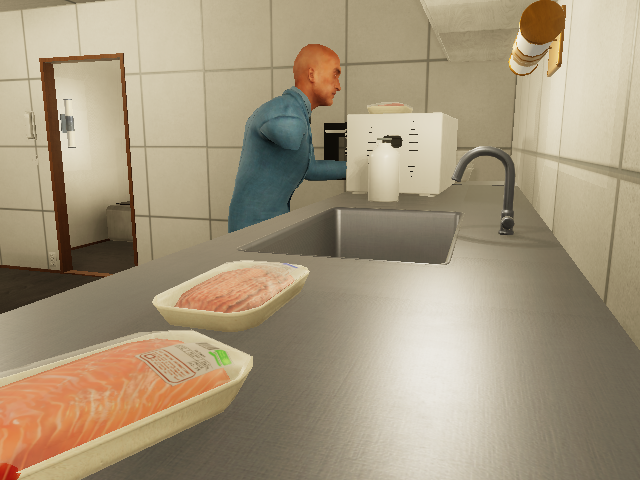} \vspace{-16pt}
        \caption{
        \scriptsize \red{$\mathrm{Inside(milk_0, microwave_0)}$}}
        \scriptsize \blue{\{$\mathrm{MoveTo(robot, microwave_0)}$, $\mathrm{PlaceIn(milk_0, microwave_0)}$\}} \vspace{-6pt}
        \label{fig:heat_milk_4}
    \end{subfigure}
    \begin{subfigure}[t]{0.196\linewidth}
        \centering
        \includegraphics[width=\linewidth, height=0.7\linewidth]{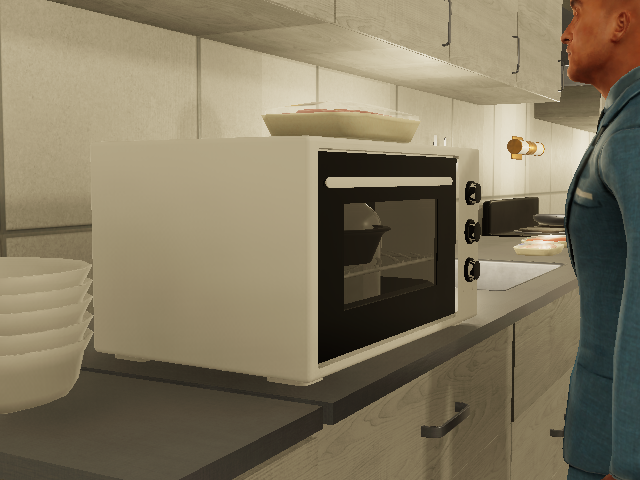} \vspace{-16pt}
        \caption{
        \scriptsize \red{$\mathrm{stateIsOn(microwave_0)}$}}
        \scriptsize \blue{\{$\mathrm{stateOn(microwave_0)}$\}} \vspace{-6pt}
        \label{fig:heat_milk_5}
    \end{subfigure}
    \caption{\modelname generalizes to unseen verbs such as \textit{heat} when only the verb \textit{boil} is seen at training.}
    \label{fig:heat_milk}
\end{figure*}
\begin{figure*}[!t]
    \centering 
    \captionsetup[subfigure]{labelformat=empty}
    \begin{subfigure}[t]{0.196\linewidth}
        \centering
        \includegraphics[width=\linewidth, height=0.7\linewidth]{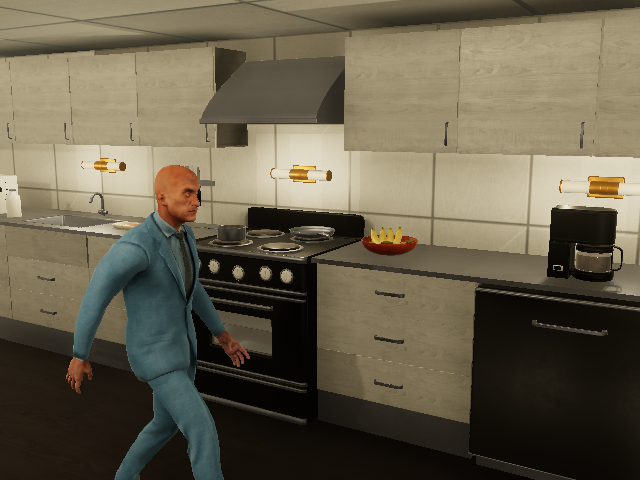} \vspace{-16pt}
        \caption{Initial State\\Instr.:\textit{``fetch milk from the fridge''}} \vspace{-6pt}
        \label{fig:get_milk_1}
    \end{subfigure}
    \begin{subfigure}[t]{0.196\linewidth}
        \centering
        \includegraphics[width=\linewidth, height=0.7\linewidth]{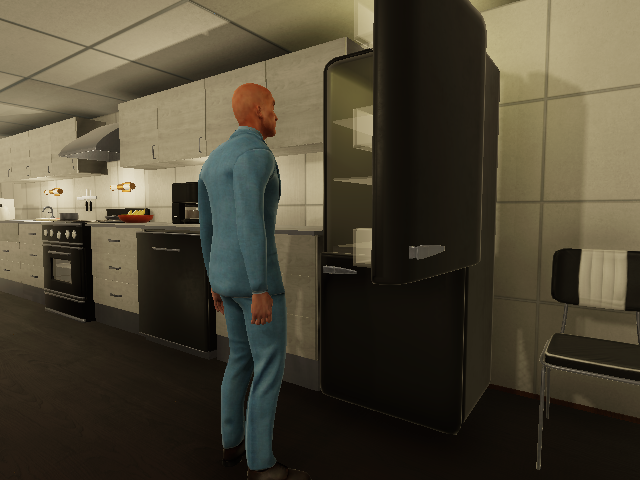} \vspace{-16pt}
        \caption{
        \scriptsize \red{$\mathrm{stateIsOpen(fridge_0)}$}}
        \scriptsize \blue{\{$\mathrm{MoveTo(robot, fridge_0)}$, $\mathrm{stateOpen(fridge_0)}$\}} \vspace{-6pt}
        \label{fig:get_milk_2}
    \end{subfigure}
    \begin{subfigure}[t]{0.196\linewidth}
        \centering
        \includegraphics[width=\linewidth, height=0.7\linewidth]{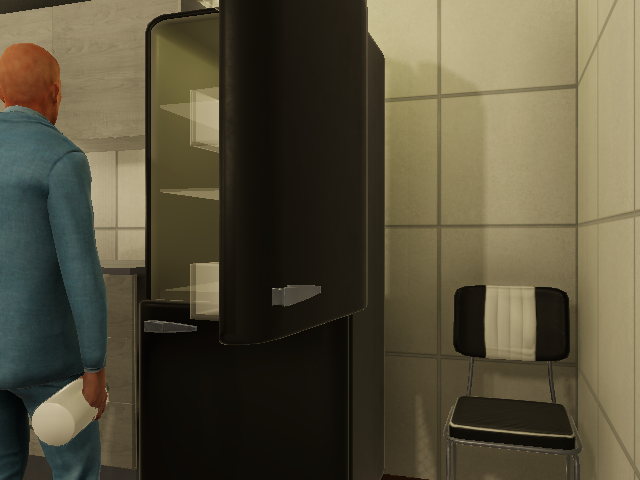} \vspace{-16pt}
        \caption{
        \scriptsize \red{$\mathrm{ConnectedTo(milk_0, robot)}$}}
        \scriptsize \blue{\{$\mathrm{MoveTo(robot, milk_0)}$, $\mathrm{Grasp(robot, milk_0)}$\}} \vspace{-6pt}
        \label{fig:get_milk_3}
    \end{subfigure}
    \begin{subfigure}[t]{0.196\linewidth}
        \centering
        \includegraphics[width=\linewidth, height=0.7\linewidth]{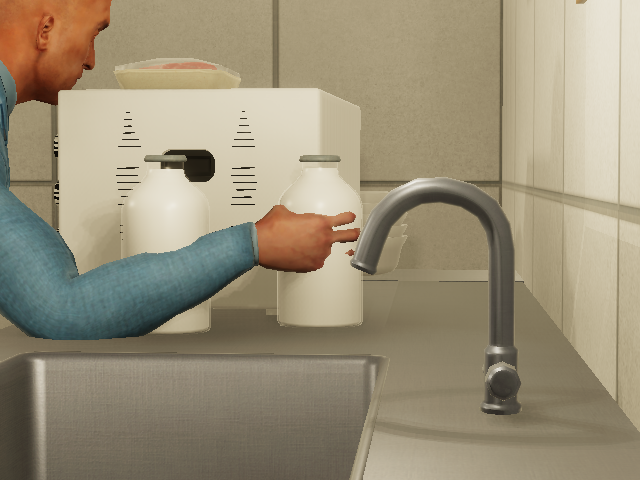} \vspace{-16pt}
        \caption{
        \scriptsize \red{$\mathrm{OnTop(milk_0, counter_0)}$}}
        \scriptsize \blue{\{$\mathrm{MoveTo(robot, counter_0)}$, $\mathrm{PlaceOn(milk_0, counter_0)}$\}} \vspace{-6pt}
        \label{fig:get_milk_4}
    \end{subfigure}
    \begin{subfigure}[t]{0.196\linewidth}
        \centering
        \includegraphics[width=\linewidth, height=0.7\linewidth]{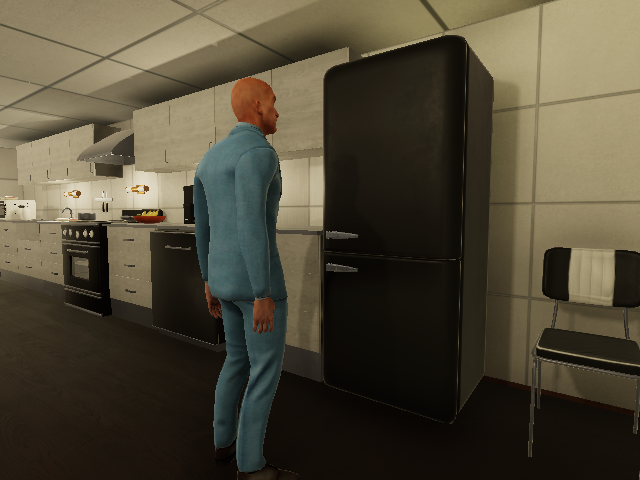} \vspace{-16pt}
        \caption{
        \scriptsize \red{$\mathrm{stateIsClose(fridge_0)}$}}
        \scriptsize \blue{\{$\mathrm{MoveTo(robot, fridge_0)}$, $\mathrm{stateClose(fridge_0)}$\}} \vspace{-6pt}
        \label{fig:get_milk_5}
    \end{subfigure}
    \caption{\modelname generalizes to unseen verbs such as \textit{fetch} when only the \textit{get} and \textit{bring} verbs are seen at training.}
    \label{fig:get_milk}
\end{figure*}

\begin{figure*}[!t]
    \centering
    \captionsetup[subfigure]{labelformat=empty}
    \begin{subfigure}[t]{0.196\linewidth}
        \centering
        \includegraphics[width=\linewidth, height=0.7\linewidth]{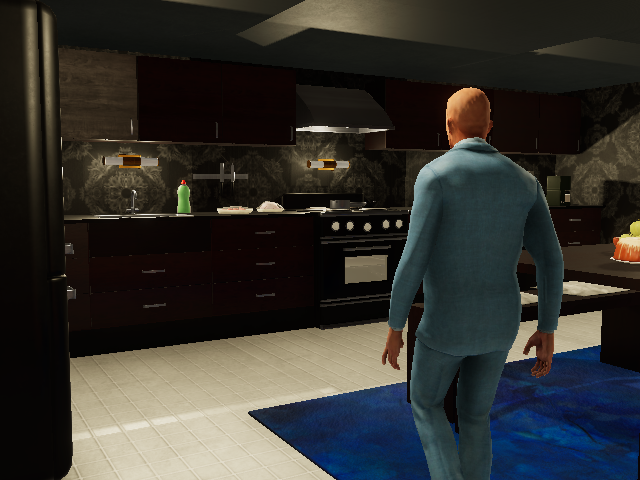} \vspace{-16pt}
        \caption{Initial State\\Instr.: \textit{``arrange pillows on the table''}} \vspace{-6pt}
        \label{fig:get_pillows_1}
    \end{subfigure}
    \begin{subfigure}[t]{0.196\linewidth}
        \centering
        \includegraphics[width=\linewidth, height=0.7\linewidth]{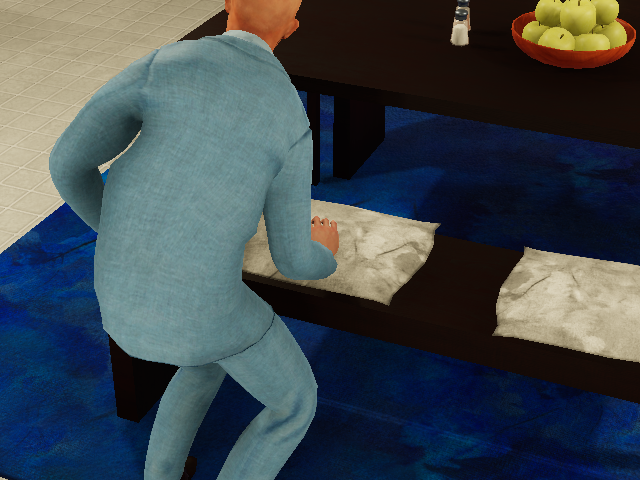} \vspace{-16pt}
        \caption{
        \scriptsize \red{$\mathrm{ConnectedTo(pillow_0, robot)}$}}
        \scriptsize \blue{\{$\mathrm{MoveTo(robot, pillow_0)}$, $\mathrm{Grasp(robot, pillow_0)}$\}} \vspace{-6pt}
        \label{fig:get_pillows_2}
    \end{subfigure}
    \begin{subfigure}[t]{0.196\linewidth}
        \centering
        \includegraphics[width=\linewidth, height=0.7\linewidth]{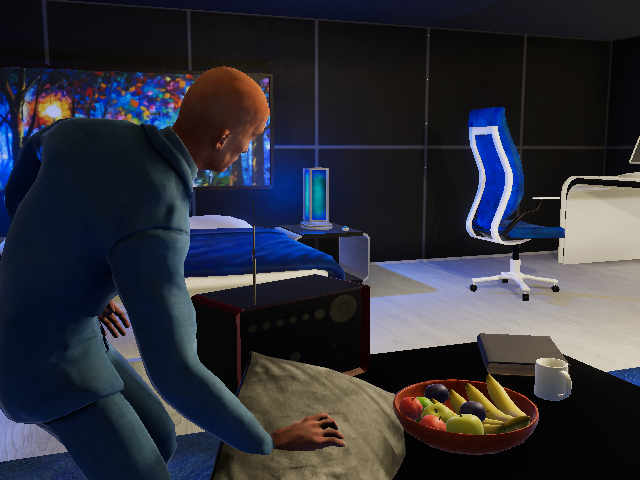} \vspace{-16pt}
        \caption{
        \scriptsize \red{$\mathrm{OnTop(pillow_0, table_0)}$}}
        \scriptsize \blue{\{$\mathrm{MoveTo(robot, table_0)}$, $\mathrm{PlaceOn(pillow_0, table_0)}$\}} \vspace{-6pt}
        \label{fig:get_pillows_3}
    \end{subfigure}
    \begin{subfigure}[t]{0.196\linewidth}
        \centering
        \includegraphics[width=\linewidth, height=0.7\linewidth]{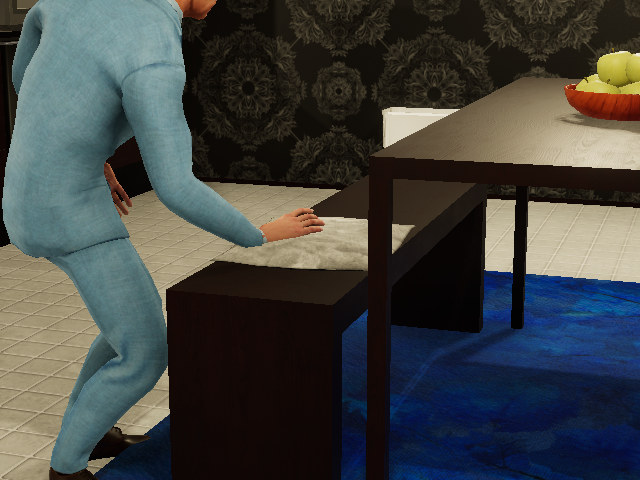} \vspace{-16pt}
        \caption{
        \scriptsize \red{$\mathrm{ConnectedTo(pillow_1, robot)}$}}
        \scriptsize \blue{\{$\mathrm{MoveTo(robot, pillow_1)}$, $\mathrm{Grasp(robot, pillow_1)}$\}} \vspace{-6pt}
        \label{fig:get_pillows_4}
    \end{subfigure}
    \begin{subfigure}[t]{0.196\linewidth}
        \centering
        \includegraphics[width=\linewidth, height=0.7\linewidth]{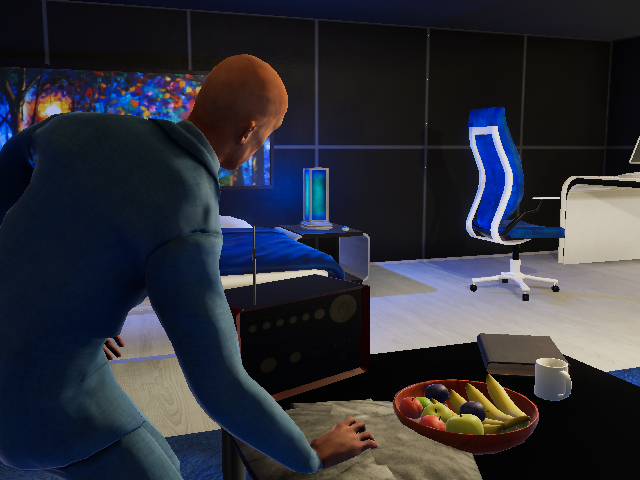} \vspace{-16pt}
        \caption{
        \scriptsize \red{$\mathrm{OnTop(pillow_1, table_0)}$}}
        \scriptsize \blue{\{$\mathrm{MoveTo(robot, table_0)}$, $\mathrm{PlaceOn(pillow_1, table_0)}$\}} \vspace{-6pt}
        \label{fig:get_pillows_5}
    \end{subfigure}
    \caption{\modelname can generalize even in case of paraphrased instructions.}
    \label{fig:get_pillows}
\end{figure*}

\section{Implementation and Training Details}
We detail the hyper-parameters for the \modelname architecture introduced in this paper.
The Parameterized ReLU activation function with a 0.25 negative input slope was used in all hidden layers of the world-state encoder described in eq.~\ref{eq:world_encoder}. The word embeddings (derived from $\mathrm{ConceptNet}$) have a size of 300. Sentence embeddings from $\mathrm{SBERT}$ have a size of 384. Additionally, the properties of objects such as $\mathrm{isGraspable}$, $\mathrm{isContainer}$, etc. is encoded as a one-hot vector of size 12. Object states such as $\mathrm{Open/Close}$ are also encoded as one-hot vectors of size 7.

\begin{itemize}
    \item \textit{World-State Encoder:} The relational information between object nodes, in the form of adjacency matrix, is encoded using a 1-layer FCN of layer size $\mathcal{O}(s_t) \times 1$ with the $\mathrm{sigmoid}$ activation function.
    \item \textit{Temporal Context Encoder:} A Long Short Term Memory (LSTM) layer of size 128 is used to encode the temporal history of predicted goal constraints, encoded as a concatenation of likelihood vectors as $[\tilde{R}^+_t, \tilde{o}^1_t, \tilde{o}^2_t]$.
    \item \textit{Instruction Conditioned Attention:} The language instruction is embedded using a pretrained SBERT and then passed through a 1-layer FCN of output size 128. We attend over the state encoding conditioned on the encoded input task instruction where the attention weights for each object in the state are generated using a 1-layer FCN with the $\mathrm{sigmoid}$ activation function. Next, we use these goal-conditioned state object embeddings to generate the encoding of the objects in the instruction using Bahdanau style self-attention. This is again achieved using a 1-layer FCN with the $\mathrm{sigmoid}$ activation function, generating attention weights for each of the objects specified in the input instruction.
    \item \textit{Goal Constraint Decoder.} After generating the final goal embedding by concatenating the instruction attended world state $\tilde{s}_t$, encoding of the constraint-history $\tilde{\eta}_t$, encoding of instruction objects $\tilde{l}_{obj}$ and the sentence encoding $\tilde{l}$, we predict the predicates. We pass $[\tilde{s}_t, \tilde{\eta}_t, \tilde{l}_{obj}, \tilde{l}]$ through a 1-layer FCN  of size 128 and $\mathrm{PReLU}$ activation function. We predict a pair of positive and negative constraints as relations $R_t^+(o^1_t, o^2_t)$ and $R_t^-(o^2_t, o^4_t)$. There are two identical decoder heads to independently predict the positive and negative constraints. To predict the relation in each constraint, a 1-layer FCN was used, with an output layer with size $\mathcal{S}$. We take the output of the Gumbel-softmax function and pass it to the decoder of the first object. The $o^1_t$ predictor generates a likelihood vector of size $\mathcal{O}(s_t)$ by passing it through a 1-layer FCN and the $\mathrm{softmax}$ activation function. The Gumbell-softmax output of the likelihood vectors of the relation and the first object are sent to the $o^2_t$ predictor to predict likelihoods for all object embeddings. This part was implemented as a 1-layer FCN with output size of $\mathcal{O}(s_t)$ followed by a $\mathrm{softmax}$ activation function. 
\end{itemize}

\noindent \textbf{Model Training.} 
We use the Adam optimizer to train our model~\citep{kingma2014adam} with a learning rate of $5 \times 10^{-4}$. We use the early stopping criterion with loss on the validation set as the signal. We also decay the learning rate by $1/5$ every 50 epochs. In training, we use a constant teacher-forcing probability of $p = 0.2$ of using the planner to iteratively update the state instead of using ground-truth state-action sequence.

\noindent \textbf{System Specifications.}
The \modelname neural network is trained and evaluated on a machine with following hardware specifications: \textit{CPU:} 2x Intel Xeon E5-2680 v3 2.5GHz/12-Core ``Haswell", \textit{GPU:} 2x NVIDIA K40 (12GB VRAM, 2880 CUDA cores), \textit{Memory:} 16GB RAM.

\section{Additional Results}

\noindent \textbf{Performance with increasing complexity.}
Figure \ref{fig:sensitivity} characterizes the variation in SJI, IED, F1 and GRR scores as the size of ground truth constraint set  (($len(\delta^+) + len(\delta^-)$) increases.  We observe graceful degradation in performance as constraint of model as the plan length increases. This suggests that there is a scope of improvement in the model to achieve good performance agnostic to plan length. However, we observe that the performance of \modelname is better than the baseline for most cases. And this performance gap between the two models widens for larger constraint sets, showing that the neural approach in \modelname is able to effectively encode the temporal context enabling it to outperform the baseline in multi-stage long-horizon tasks.

\noindent \textbf{Generalization.}
Figures \ref{fig:heat_milk}, \ref{fig:get_milk}, and \ref{fig:get_pillows} demonstrate the ability of \modelname to generalize to language instructions unseen during training. These instructions correspond to verb frames that are \textit{out-of-distribution} from the training data. Robust inference of conjunctive goal predicates enables the symbolic planner to generate feasible plans and reach goal states for these unseen tasks. Figure \ref{fig:heat_milk} shows a trace of the inferred goal-predicates and actions generated by the \textsc{Rintanen} planner for an input language instruction of \textit{``boiling the milk''} when the instruction uses the verb \textit{heat} instead of \textit{boil} in the training data. \modelname correctly predicts the predicate of opening microwave, placing the bottle of milk inside and turning on the microwave to heat the milk. Similarly, Figure \ref{fig:get_milk} shows a trace of predicates and actions in case the input language instruction is \textit{``fetch milk from the fridge"} when the instruction has the unseen verb \textit{fetch} instead of known verbs such as \textit{get} or \textit{bring}. Figure \ref{fig:get_pillows} shows another example where the model generalized to paraphrased sentences by performing the task correctly changing \textit{``bring pillows to the table''} to \textit{``arrange pillows on the table''}.


\end{document}